\documentclass[12pt]{article}
\usepackage{amsmath}
\usepackage{times}
\usepackage{epsf,multirow}
\usepackage{amssymb}
\usepackage{graphics}
\usepackage{latexsym}
\usepackage{amsmath}
\usepackage{xspace}
\usepackage{amssymb}
\usepackage{amsmath}
\usepackage{multirow}
\usepackage{graphicx}
\usepackage{subfigure}
\usepackage{makeidx}
\usepackage{multirow}
\usepackage{url}
\usepackage{alltt}
\usepackage{setspace}
\usepackage{csquotes}
\usepackage{enumitem}
\usepackage{moreverb}    

\newcommand*{\nc}[2]{#1\mathbin{\left| \sim \vphantom{#1#2} \right.}#2}%

\begin{document} \sloppy
\pagenumbering{roman}
\setcounter{page}{1}
\thispagestyle{empty}

\begin{center}

\vspace{1in}
{\Large\bf Controlled Natural Languages and Default Reasoning }

\vspace{0.2in}
by\\
\vspace{0.05in}
 {\Large\it Tiantian Gao}\\

\vspace{0.2in}
Department of Computer Science

Stony Brook University

\end{center}

\begin{abstract}

Controlled natural languages (CNLs) are effective languages for knowledge representation and reasoning. They are designed based on certain natural languages with restricted lexicon and grammar. CNLs are unambiguous and simple as opposed to their base languages. They preserve the expressiveness and coherence of natural languages. In this report, we focus on a class of CNLs, called machine-oriented CNLs, which have well-defined semantics that can be deterministically translated into formal languages, such as Prolog, to do logical reasoning. Over the past 20 years, a number of machine-oriented CNLs emerged and have been used in many application domains for problem solving and question answering. However, few of them support nonmonotonic inference. In our work, we propose nonmonotonic extensions of CNL to support defeasible reasoning.

In the first part of this report, we survey CNLs and compare three influential systems: Attempto Controlled English (ACE), Processable English (PENG), and Computer-processable English (CPL). We compare their language design, semantic interpretations, and reasoning services. In the second part of this report, we first identify typical nonmonotonicity in natural languages, such as \textit{defaults}, \textit{exceptions} and \textit{conversational implicatures}. Then, we propose their representation in CNL and the corresponding formalizations in a form of defeasible reasoning known as \textit{Logic\ Programming\ with\ Defaults\ and\ Argumentation\ Theory} (LPDA).

\end{abstract}

\newpage
\pagestyle{plain}
\tableofcontents

\newpage


\pagenumbering{arabic}
\setcounter{page}{1}


\newpage

\section{Introduction}

Controlled natural languages (CNLs) are effective languages for knowledge representation and reasoning. According to Kuhn, ``A controlled natural language  is a constructed language that is based on a certain natural language, being more restrictive concerning lexicon, syntax, and/or semantics while preserving most of its natural properties'' \cite{kuhn2013survey}. Unlike the languages that develop naturally, constructed languages are the languages whose lexicon and syntax are designed with intent. A CNL is constructed on the basis of an existing natural language, such as English, French, or German. Words in the lexicon of a CNL mainly come from its base language. Some CNLs may include special symbols in their lexicon. For example, in Common Logic Controlled English (CLCE) the parentheses are used for deeply nested sentences and for lists of more than two elements \cite{sowa2004common}. Words may or may not be used in the same manner as in the base language. Some words are used with fewer senses or reserved as key-words for specific purposes. In CLCE, the word ``every`` is treated as a universal quantifier. It does not allow any universally quantified noun phrases to be used as the object of a preposition. CNLs have a well-defined syntax to form phrases, sentences and texts. The syntax of a CNL is generally simpler than that of the source language. Sentences are interpreted in a deterministic way. CNLs are more accurate than natural languages, because the language is more restrictive, but not all CNLs have formal semantics. Those that have formal semantics can be processed by computers for knowledge representation, machine translation, and logical reasoning. Although a CNL may deviate from its base language in the lexicon, syntax, and/or semantics, it still preserves most of the natural properties of the base language, so the reader would correctly comprehend the CNL with little effort.

CNLs generally fall into two categories: human-oriented CNLs and machine-oriented CNLs \cite{schwitter2010controlled}. Human-oriented CNLs are designed to make the texts easier for readers to understand. They are applied in technical writings and inter-person communications. Basic English was the first English-based controlled natural language created by Charles Ogden in 1930 \cite{ogden1944basic}. It has a tremendous impact on the development of controlled natural languages. Basic English has 850 core root words which are composed of 600 nouns, 150 adjectives and 100 functional words that put the nouns and adjectives into operation. These 850 words can do all the work that 20,000 English words do in daily life. Root words can be extended to form plurals, negative adjectives, etc.. For example, plurals are formed by appending an ``S`` to the end of a root word. An adjective is given a negative meaning with the prefix, ``UN-``. Basic English substitutes verbs in English with operators. The operator is one of the 18 verbs: put, take, give, get, come, go, make, keep, let, do, be, seem, have, may, will, say, see, and send. A verb can be described by an operator combined with a preposition or a noun. For instance, Basic English uses “put in” to represent the verb, “inject”, in English. There are 10 rules of grammar that define the order of words in a sentence, how root words can be extended to form plurals, adjectives, adverbs, negative adjectives, and compound words, the composition of questions, and so on. In addition to the 850 core root words, Basic English also defines a list of international words. The international words are intended for scientific and technical writing. Basic English is used in business, science, economics, and politics. The benefits of Basic English are two-fold. First, it can serve as an international auxiliary language and work as an aid for teaching English as a second language. Second, it provides a simple introduction to English for foreign language speakers. 

Other influential human-oriented CNLs include Special English and Simplified Technical English (STE). Special English was developed by Voice of America in 1959 \cite{voice2007voa}. The vocabulary is limited to 1580 words. Special English has been used in a multitude of daily news programs in the United States. It also serves as a resource for English learners and has been adopted by other countries for news broadcasting. Simplified Technical English (STE) was developed for aerospace industry maintenance manuals \cite{asd2007asd}. The dictionary consists of approved and unapproved words. Unapproved words are not allowed to be used. Instead, the dictionary provides alternative words that refer to the same meaning. STE also allows users to add approved words to the dictionary. STE, as a subset of English, reduces ambiguities and improves clarity and comprehensibility through restrictions on grammar and style. It also helps in computer-assisted and machine translation.

Machine-oriented CNLs, as opposed to human-oriented ones, have formal semantics which can be understood and processed by computers for the purpose of knowledge representation and logical reasoning. Attempto Controlled English (ACE) was the first CNL that can be translated to first-order logic \cite{fuchs2008attempto}. ACE is a subset of English defined by a restricted grammar along with interpretation rules that control the semantic analysis of grammatically correct ACE sentences. ACE uses discourse resolution structure (DRS) as the logical structure to represent the semantics of a set of ACE sentences \cite{kamp1993discourse}. ACE is supported by a language processor, Attempto Parsing Engine (APE), and a reasoner, RACE. APE is an online language processor that allows users to compose ACE sentences as input and generates their semantics in DRS and first-order logic clauses as output. RACE is a CNL reasoner that supports theorem proving, consistency checking, and question answering. ACE has been applied to various areas such as semantic web, bioinformatics and so on. 

Other examples include PENG, CELT and CLP \cite{schwitter2002english,pease2010controlled,clark2005acquiring}. Both PENG and CELT are inspired by ACE. They are a subset of English with restricted grammar and use DRS as semantic representation. Unlike ACE, PENG does not require users to learn the grammar of the language. Instead, it designs a predictive editor that informs users of the look-ahead information that guides users to proceed based on the structure of the current sentence. CELT translates the controlled language to formal logical statements that use terms in an existing large ontology, the Suggested Upper Merged Ontology (SUMO) \cite{niles2001origins}. This gives each term multiple definitions. CPL is a CNL developed by Boeing \cite{clark2005acquiring}. It uses Knowledge Machine (KM), an advanced frame-based language, to represent the semantics of its language \cite{Clark99km-}. The translation is through three main steps: parsing, generation of an intermediate “logical form” (LF), and conversion of the LF to statements in the KM knowledge representation language. The KM statements can be used for reasoning and question-answering. 

The above four machine-oriented CNLs are designed for general purposes. They can be applied in various application domains. Some machine-oriented CNLs are designed for specific application domains. For example, SBVR Structured English is intended for describing business vocabulary and for representing business rules \cite{chapin2005semantics}. Rabbit is a CNL that can be translated into OWL and achieves both comprehension by domain experts and computational experts \cite{hart2008rabbit}. LegalRuleML is used for representing legislative documents in XML-based rules in order to conduct legal reasoning \cite{palmirani2011legalruleml}. 

Although a number of machine-oriented CNLs exist, few of them support nonmonotonic reasoning. In our work, we propose nonmonotonic extensions of CNL to support defeasible reasoning. A logic is nonmonotonic if some conclusions can be defeated by the addition of new knowledge. This contrasts with monotonic formalisms, where the addition of new knowledge will not invalidate any previously derived conclusion. Nonmonotonic reasoning represents the natural way of human reasoning in real life. People draw conclusions based on a number of unstated default assumptions which are supposed to be true. Some conclusions may be retracted when the addition of new knowledge violates one of the default assumptions. For instance, given the following knowledge base,

\begin{enumerate}
  \item Typically, birds fly.
  \item Penguins are birds.
  \item Penguins do not fly.
  \item Tweety is a bird.
\end{enumerate}
\noindent
the first sentence is in fact interpreted as ``in the absence of any information to the contrary, we assume that birds fly.'' Hence, it is reasonable to make the conclusion that ``Tweety flies.'' However, if later we add the fact that ``Tweety is a penguin,'' we will withdraw the previous conclusion and conclude that ``Tweety does not fly.'' Classical approaches that deal with \textit{defaults} include default logic \cite{reiter1980logic}, circumscription \cite{mccarthy1980circumscription}, and autoepistemic logic \cite{marek1991autoepistemic}. Modern approaches include answer set programming (ASP) \cite{marek1991autoepistemic} and logic programming with defaults and argumentation theories (LPDA) \cite{wan2009logic}.  In this report, we focus on three logical frameworks: circumscription, ASP and LPDA, where circumscription is used in Wainer's approach to model \textit{conversational\ implicatures}, ASP is used in PENG to handle \textit{defaults} and \textit{exceptions}, and LDPA is used in our approach to model nonmonotonicity in CNL. They will be discussed in Section 5.2, 3.4, and 5.1 respectively. Technical details of these three frameworks can  be found in the appendix.

The following is organized as follows: in Section 2-4, we study three influential CNL systems, Attempto Controlled English (ACE), Processable English (PENG), and Computer-processable English (CPL). We compare their language design, semantic interpretations, and reasoning services. In Section 5, we first identify typical nonmonotonicity in natural languages, such as \textit{defaults}, \textit{exceptions} and \textit{conversational implicatures}. Then, we propose their representation in CNL and the corresponding formalizations in a form of defeasible reasoning known as \textit{Logic\ Programming\ with\ Defaults\ and\ Argumentation\ Theory} (LPDA).

\section{Attempto Controlled English}

Attempto Controlled English (ACE) will be discussed in this section. Subsection 1-4 introduce the language properties of ACE, including \textit{vocabulary}, construction rules and interpretation rules. Subsection 5 describes Discourse  Representation Structure (DRS), the semantic interpretation of ACE sentences. Subsection 6 shows the Attempto Parsing Engine (APE). Subsection 7 presents the Attempto reasoning engine, RACE reasoner.

\subsection{Vocabulary}

The \textit{vocabulary} consists of function words, such as determiners, articles, pronouns, and quantifiers, some fixed phrases (e.g. ``there is a'' and ``it is not the case that''), and 100,000 content words, including adverbs, adjectives, nouns, verbs, and prepositions. ACE also allows users to add new content words to the lexicon. Content words are written as Prolog atoms. The predicates describe the part of speech (POS) of the words. Each predicate has at least two arguments. The first argument is a word in ACE lexicon. The second argument specifies the corresponding logical symbol of the first argument used in representing the semantics of ACE sentences. A predicate may represent some additional information by adding a few more arguments. Below is an example that describes the Prolog representations of the words ``fast,'' ``faster,'' and ``fastest'' respectively,

\begin{enumerate}
  \item $adv(fast, fast)$.
  \item $adv\_comp(faster, fast)$.
  \item $adv\_sup(fastest, fast)$.
\end{enumerate}
\noindent
where predicate $adv$ says that ``fast'' is an adverb, predicate $adv\_comp$ says that ``faster'' is a comparative adverb, predicate $adv\_sup$ says that ``fastest'' is a superlative adverb. All three words use the same logical symbol, ``fast,'' in logical representations. 

\subsection{Construction Rules}
 
The \textit{construction\ rules} define the format of words, phrases, sentences, and ACE texts. 
\begin{description}
  \item[Words] Function words and some fixed phrases are not allowed to be modified by users. Users can create compound words by concatenating two or more content words with hyphens.
  \item[Phrases] Phrases include noun phrases, modifying nouns, modifying noun phrases, verb phrases, and modifying verb phrases. Noun phrases in ACE are a subset of noun phrases in English plus arithmetic expressions, sets, and lists. Modifying nouns and noun phrases are those that are proceeded or followed by adjectives, relative clauses, or possessives. Verb phrases in ACE form a subset of verb phrases in English with specific definitions of negations and modalities. Modifying verb phrases are those that are accompanied by adverbs or preposition phrases.  
  \item[Sentences] ACE sentences include declarative, interrogative, and imperative sentences. Declarative sentences include simple sentences, there is/are-sentences, boolean formulae, and composite sentences. Interrogative sentences include yes/no queries, wh-queries, and how much/many-queries which end with a question mark. Imperative sentences are commands and end with an exclamation mark. 
  \item[ACE Texts] ACE texts are sequences of declarative, interrogative, and imperative sentences. 
\end{description}
 
\subsection{Interpretation Rules}

ACE restricts sentences to be interpreted in only one deterministic way. The interpretation rules specify how a grammatically correct ACE sentence is translated. Below are three examples of the interpretation rules:

\begin{enumerate}
  \item Prepositional phrases modify the verb not the noun
  \begin{itemize}
    \item A customer \{enters a card with a code\} .
  \end{itemize}
  \item Relative clauses modify the immediately preceding noun
  \begin{itemize}
    \item A customer enters \{a card that has a code\} .
  \end{itemize}
  \item Anaphora is resolved using the nearest antecedent noun that agrees in gender and number
  \begin{itemize}
    \item Brad was born in Seattle. It's a beautiful place.
  \end{itemize}
\end{enumerate}

The first sentence can be paraphrased in two ways in English,  ``A customer uses a code to enter a card'' and ``A customer enters a card. The card has a code.'' In ACE, it is only translated in the first way. In the second sentence, given a relative clause, an ACE sentence only refers to the closest noun that precedes it. Hence, the relative clause, ``that has a code,'' modifies the noun, ``card.''  In the third sentence, ACE resolves the pronoun, ``it,'' in the second sentence by looking for the nearest antecedent noun that agrees with gender and number, so ACE associates ``it'' with the word ``Seattle.''

\subsection{Evaluation of ACE}

There are advantages and disadvantages in the design of ACE. The language has a large vocabulary and a multitude of construction rules. The number of words in the ACE lexicon is more than half of the words (171,476) that are used in current English according to the second edition of the 20-volume Oxford English Dictionary \cite{simpson1989oxford}.  Its construction rules cover various sentence structures plus mathematical expressions and boolean formulae. On one hand, these allow users to describe more things and thereby make the language very expressive. On the other, these add complexity to the language. It is difficult for casual users to learn the language and takes them many trials to get to a grammatically correct sentence. Interpretation rules ensure a deterministic way to interpret an ACE sentence and thus avoid many ambiguities  that exist in English, however this does not always lead to natural interpretations and often results in stilted English. For example, ``Brad is an actor. He is handsome.'' The pronoun, ``He,'' in the second sentence refers to the proper name, ``Brad,'' in the first sentence. However, ACE revolves anaphora by referring ``He'' back to the noun, ``actor.'' Users can only avoid this problem by rephrasing the sentences in a way that follows the ACE interpretation rules. 

\subsection{Semantic Interpretations}

The semantics of ACE are represented by Discourse Resolution Structures (DRSs) \cite{kamp1993discourse}. A DRS is a diagram structure consisting of two parts:
\begin{itemize}
  \item a set of discourse referents (discourse variables), called the ``universe'' of the DRS, which will always be displayed at the top of the diagram \cite{kamp1993discourse}.
  \item a set of DRS-conditions, typically displayed below the universe \cite{kamp1993discourse}.
\end{itemize}

A referent can stand for an object introduced by a noun, a predicate introduced by a verb, etc.. A condition can be either simple or complex. A simple condition is a logical atom followed by an index. A complex condition is constructed from other DRSs connected by operators such as negation, disjunction, and implication. For each simple condition, the logical atom is restricted to one of the 8 predicates: object, property, relation, predicate, modifier\textunderscore adv, modifier\textunderscore pp, has\textunderscore part, and query. Their definitions are given in \cite{ifi-2009.04}. The index shows the location of the input from which the condition is introduced.

\begin{figure}[h]
    \centering
    \includegraphics[scale=1]{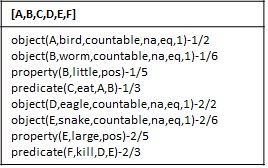}
    \caption{DRS for ``A bird eats a little worm. An eagle kills a large snake.''}
    \label{drs-1}
\end{figure}

The two ACE sentences, ``A bird eats a little worm. An eagle kills a large snake,'' are represented in one logical unit, as shown in Figure \ref{drs-1}. In the first condition, the object-predicate represents the bird-object. It has 6 arguments: A, bird, countable, na, eq, and 1. The first argument, ``A,'' is the label for the bird-object. It can be referenced by other conditions if the conditions are within the scope of the DRS to which A belongs. The second argument indicates that the word, ``bird,'' introduces the object-predicate. The third argument identifies the bird-object as a countable object. The fourth argument shows that the bird-object does not come together with any measurement unit (e.g. kg, cm) in the input. The fifth and the sixth arguments specify the quantity of the bird-object to be one. The index, ``1/2,'' signifies that the bird-object is introduced by the second word of the first ACE sentence in the input. The same applies to the worm-, eagle-, and snake-objects. In the third condition, the property-predicate describes the little-property. It has 3 arguments: B, little, and pos. The first argument, ``B,'' refers to the worm-object. The second argument shows that the word ``little'' introduces the little-property. The third argument denotes the degree of the little-property as positive. The fourth condition represents the eat-predicate. It has 4 arguments: C, eat, A, and B. The first argument, ``C,'' is the label for the eat-predicate. Similar to the bird-object, C can be used by other conditions as well. The second argument indicates that this predicate introduces the word ``eat.'' The third and fourth arguments refer to the bird- and worm-objects. 

DRSs can be nested. DRS-conditions are allowed to use the referents from their ancestor but not descendant DRSs. A multi-sentential paragraph is always represented by one DRS instead of a conjunction of individual DRSs. This is because sentences may be cross-referenced among each other. Representing a paragraph by a conjunction of individual DRSs cannot indicate their connections. A DRS is constructed incrementally by processing each sentence in order. Once a sentence gets processed, new referents and conditions are incorporated into the current DRS, which is constructed from all preceding sentences. Anaphora is resolved by referring to a variable from the current DRS. 

\subsection{Attempto Parsing Engine}
Attempto parsing engine (APE) is a language processor that accepts ACE texts as input and generates paraphrases, DRSs and first-order logic clauses. Paraphrases indicate to users how APE comprehends the ACE sentences. Users can rephrase their input to get another paraphrases if they do not accept APE's interpretations. APE generates first-order logic clauses from DRSs using the methods introduced in \cite{kamp1993discourse}. The core part of APE is a top-down parser that implements the construction and interpretation rules. The parser uses the Definite Clause Grammar (DCG) enhanced by feature structures \cite{fuchs1995specifying}. It is written in GULP (Graph Unification Logic Programming) which extends Prolog by adding the operators ``:'' and ``..,'' and a number of built-in predicates \cite{covington1994gulp}. The operator ``:'' binds a feature name to its value. The operator ``..'' joins one feature-value pair to the next. In XSB Prolog, list can do the same thing as ``..'' \cite{Sagonas94xsbas}.

\begin{figure}[h]
\hfill
\subfigure[DRS]{\includegraphics{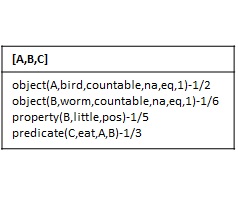}}
\hfill
\subfigure[First-order Logic]{\includegraphics{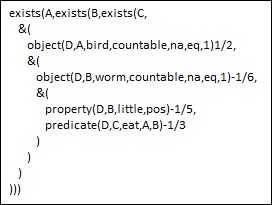}}
\hfill
\caption{The ACE sentence, ``A bird eats a little worm.''}
\label{drs-2}
\end{figure}

There are several problems with APE. First, logical clauses are not represented in the usual sense of first-order logic. An example is shown in Figure \ref{drs-2}. In Attempto, a noun is represented by a pre-defined predicate \textit{object} where the fifth and the sixth fields denote the quantity information. Here, the predicate $object(A,bird,countable,na,eq,1)$ indicates that there is one bird. If the sentence is changed to ``Two birds eat a little worm,'' then the \textit{object} predicate for bird becomes $object(A,bird,countable,na,eq,2)$. The way Attempto does simply treats the quantification information as an adjective modifier to the noun. If there is a query that asks for the quantity information of bird, it will check the fifth and the sixth fields of the \textit{object} predicate. This is not represented correctly in first-order logic. In first-order logic, the  quantity information should be represented by existential quantifiers. For instance, the sentence, ``There are two birds,'' is translated into first-order logic as shown below:
\begin{equation*}
\exists x_{1} \exists x_{2} ( bird(x_{1}) \wedge bird(x_{2}))
\end{equation*}
\noindent
where $x_{1}$ and $x_{2}$ denote two unique bird entities.

\begin{figure}[h]
\hfill
\subfigure[DRS]{\includegraphics{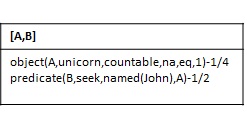}}
\hfill
\subfigure[First-order Logic]{\includegraphics{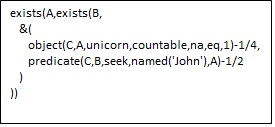}}
\hfill
\caption{The ACE sentence, ``John seeks a unicorn.''}
\label{drs-4}
\end{figure}

Second, APE does not recognize certain sentence structures such as ``he seeks...'' which cannot be represented in first-order logic. For example, Figure \ref{drs-4} shows the DRS and first-order logic clause for the sentence, ``John seeks a unicorn.'' In fact, ``John seeks a unicorn'' indicates that there may or may not exist a unicorn, so the unicorn cannot be simply existentially quantified as in first-order logic. This type of sentences is represented by intentional logic, which is an extension of first-order logic \cite{droste1991linguistic}. 

\subsection{RACE Reasoner}
RACE is an ACE reasoner that accepts ACE texts as input and supports consistency checking, theorem proving, and question answering \cite{fuchs2003reasoning}. RACE is implemented in Prolog. It is an extension of Satchmo, which is a theorem prover based on the model generation paradigm \cite{manthey1988satchmo}. Satchmo works with first-order logic clauses of the form $Body \rightarrow Head$ where $Body$ is true or a conjunction of logical atoms, and $Head$ is fail or a disjunction of logical atoms. Negation is expressed as an implication to false. Satchmo executes the clauses by forward-reasoning and generates a minimal finite model of clauses. RACE extends Satchmo by giving a justification for every proof, finding all minimal unsatisfiable subsets of clauses if the axioms are not consistent, etc..

\begin{figure}[h]
    \centering
    \includegraphics[scale=1]{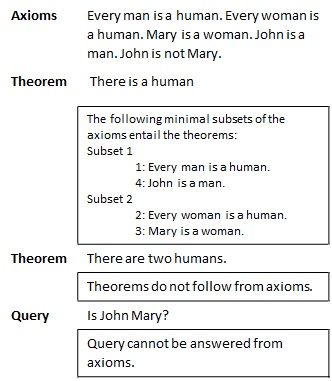}
    \caption{An example of RACE reasoning}
    \label{drs-5}
\end{figure}

RACE has certain limitations. First, RACE does not work properly in some cases. An example is shown in Figure \ref{drs-5}. There are five consistent axioms. RACE can prove that there is a human based on either the first and fourth sentences or the second and third sentences. However, RACE cannot prove that there are two humans in total. The fifth axiom says that ``John is not Mary,'' but RACE cannot answer the query, ``Is John Mary?'' from the axioms. 

Second, RACE does not support defeasible reasoning \cite{nute1994defeasible}. For example, given the sentences, ``Birds fly. Penguins are birds. Penguins don't fly,'' RACE shows that these axioms are inconsistent. This is because the second and the third axioms, which state that penguins are the exceptions of birds that fly, challenge the first axiom. Such exceptions are very common in natural languages, especially in law, legislation, policies, but they cause contradiction in RACE. 



\section{Processable English}

In this section, we introduce Processable English (PENG), a system developed by Rolf Schwitter at Macquarie University who was also one of the Attempto group members \cite{trentelman2009processable}. The first subsection gives an overview of PENG's language properties and compares it with Attempto. The second subsection presents the basics of chart parsing and its application to PENG's language processor. The third subsection discusses the reasoning service in PENG. The fourth subsection describes the way PENG handles defaults and exceptions in CNL and its support for non-monotonic reasoning based on the answer set programming (ASP) paradigm. 

\subsection{Basics of PENG}

The dictionary of PENG consists of predefined function words (including determiners, cardinals, connectives, prepositions), 3,000 content words (e.g. nouns, proper nouns, verbs, adjectives, and adverbs), and illegal words. Illegal words are those that are banned by PENG, such as wish, can, could, should, might, and all personal pronouns. Users can expand the lexicon as in ACE. 

A simple PENG sentence is constructed by the following grammar, where curly brackets indicate the enclosed elements, are optional.

\begin{quote}
Sentence $\rightarrow$  Subject + Predicate

Subject   $\rightarrow$  Determiner \{+Pre-nominal Modifier\}

\qquad  \qquad \qquad + Nominal Head  \{+Post-nominal Modifier\} $|$ Nominal Head

Predicate $\rightarrow$  \{Negation\} + Verbal Head + Complement \{+Adjunct\} 
\end{quote}

Complex sentences are built from simple sentences using coordinators (and, or), subordinators (if, before, after, while), and constructors (for every, there). PENG resolves anaphora by referring back to the most recent noun phrase that agrees in gender and number. In this way, PENG functions similarly to ACE. In addition, PENG identifies synonyms within sentences. For example, given the sentence, ``The fog hangs over Dreadsbury Mansion. The mist is creepy,''  PENG recognizes that ``mist'' is a synonym for ``fog,'' and the noun phrase ``The mist'' is an anaphora for ``The fog'' \cite{trentelman2009processable}. 

Like ACE, PENG interprets sentences in a deterministic way. Once sentences are evaluated, the PENG language processor generates paraphrases for users. If PENG misinterprets a user's intentions, it allows that user to rephrase the submission. As discussed in section 2.5, ACE uses DRSs to represent the semantics of the language; PENG also uses DRSs as its semantic representations and translates DRSs to first-order logic clauses. 

PENG has several key advantages over ACE. For one, PENG has a much smaller lexicon and simpler grammar. This makes the language very accessible and easy to learn. Most importantly, users do not need to consult the grammar manual when composing sentences. Instead, PENG can inform users of all possible words which can follow their current inputs \cite{schwitter2003ecole}. Details of this feature are discussed next.

\subsection{Chart Parsing in PENG}

The PENG language processor is based on a top-down incremental chart parser that operates on DCG grammar enhanced with feature structures. A chart parser is a parser that uses a chart to save the information of the substrings that have already been successfully analyzed. It reuses this information later if needed \cite{trentelman2009processable}. This eliminates backtracking and avoids re-discovering the same substring over and over again. 

For example, given the following context-free grammar and the sentence, ``Tom walks slowly,'' a backtracking DCG parser will start from the first rule and evaluate the non-terminal symbols, $N$ and $Vb$. After it finds the first rule fails, it will explore the second rule and re-evaluate $N$ and $Vb$. A chart parser, on the other hand, avoids this duplication of work by storing the computation results of the first rule in a chart and reusing them when exploring the second rule. 

\begin{quote}
$S \rightarrow  N$ $Vb$\\
$S \rightarrow  N$ $Vb$ $Adv$\\
$N \rightarrow Tom$\\
$Vb \rightarrow walks$\\
$Adv \rightarrow slowly$
\end{quote}

\begin{figure}[h]
    \centering
    \includegraphics[scale=1]{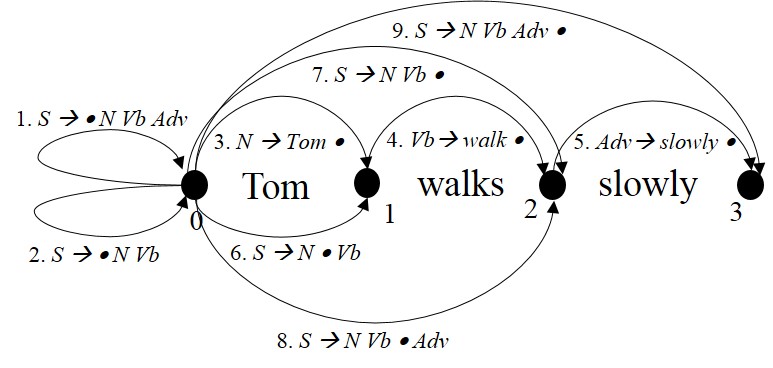}
    \caption{The chart for the sentence, ``Tom walks slowly.''}
    \label{peng-1}
\end{figure}

The chart for the above example is shown in Figure \ref{peng-1}. It is represented by a directed graph consisting of 4 vertices ($v_{0}$, $v_{1}$, $v_{2}$, $v_{3}$) and a number of edges labelled by dotted rules. A dotted rule is the same as a production rule for the context-free grammar except that it has a dot inserted somewhere in its right-hand side. The dot indicates the extent to which the hypothesis that the rule is applicable has been verified. Edges can be \textit{active} or \textit{inactive}. \textit{Active edges} refer to partially verified hypotheses; \textit{inactive edges} signify fully confirmed hypotheses. For the purposes of this explanation, edge 1 will be referred to as $e_{1}$, edge 2 to $e_{2}$, etc.. In Figure \ref{peng-1}, $e_{1}$ is an active edge that denotes a hypothesis that $S$ can be transformed to a substring that represents a $N$ $Vb$ $Adv$ sequence. $e_{8}$ is also an active edge that represents a similar hypothesis, but in this case the hypothesis has been partially verified. The parser has confirmed that $N$ $Vb$ can derive the substring ``Tom walks,'' but hasn't analyzed $Adv$ yet. $e_{3}$ is an inactive edge which indicates a fully confirmed hypothesis that $N$ can generate ``Tom.'' 

Chart parsing can be conducted either top-down or bottom-up. PENG implements the top-down approach. The parsing process involves combining active edges with inactive edges based on the fundamental rule described in \cite{trentelman2009processable}. It uses a data structure, called agenda, to maintain a list of active edges and organize the order in which these edges are executed. In PENG's approach, the agenda behaves like a stack that manages the edges in the last-in-first-out manner. In this example, the agenda is initialized with $e_{1}$ and $e_{2}$; the chart is set up with $e_{3}$, $e_{4}$, and $e_{5}$. Each time, the parser pops the first edge from the agenda and adds it to the chart. This active edge is then combined with applicable inactive edges in the chart to form new edges, which may be active or inactive. The parser keeps the inactive edges in the chart and pushes all active edges to the agenda. This process repeats until the parser finds an inactive edge that spans the whole sentence.

Incremental chart parsing is an extension of the chart parsing framework. It can handle modifications to previously parsed strings without the need for from scratch re-processing. Basically, the parser uses the edge dependency information to determine which edges to recompute. An edge is dependent on another if it is triggered by the latter. Once an edge is changed, only the edges that are dependent on it are updated. For the previous example which is shown in Figure  \ref{peng-1}, if the user deletes the last word, ``slowly,'' from the sentence, the parser will first remove $e_{5}$ and $v_{3}$. Next, the parser will eliminate $e_{9}$ because this edge is dependent on $e_{5}$. All other edges are unaffected. The incremental chart parser supports three edit operations: insertion, deletion, and replacement. Users can insert, delete, or replace a word from the sentence. Details of the corresponding algorithms are given in \cite{schwitter2003incremental}.

\subsection{Reasoning in PENG}
The PENG reasoner accepts grammatically correct sentences as input and generates the output in controlled languages as well. It supports consistency checking, informativity checking, and question/answering. Consistency checking ensures that a set of PENG sentences is semantically consistent. As compared to RACE, which is the Attempto reasoner discussed in section 2.7, PENG can also handle synonyms in sentences. For example, given the input, ``Mary lives in Seattle. Mary does not reside in Seattle,'' RACE will not find the inconsistency because it treats ``live'' and ``reside'' as words having different meanings. PENG, on the other hand, can identify the synonyms information and therefore evaluates the input as inconsistent. Informativity checking guarantees that sentences are concise with no redundant information. For instance, given the sentence, ``Mary lives in Seattle. Mary resides in Seattle,'' the second sentence violates the informativity constraint because it expresses the same meaning as the first one. For question/answering, the PENG reasoner operates in the same way as RACE. 

The implementation of the reasoner is based on a theorem-prover Otter and a model builder MACE \cite{mccune1994otter,mccune2001mace}. Otter (Organized Techniques for Theorem-proving and Effective Research) is a resolution-based theorem-prover that works on first-order logic with equality. It proves a first-order formula $\Phi$ valid by assuming $\neg \Phi$ and generating an empty clause (a contradiction) based on the inference rules (e.g. binary resolution, hyper-resolution, UR-resolution, binary-paramodulation). The PENG reasoner does not require users to specify which inference rules to use in the process of theorem proving. Instead, it relies on Otter's autonomous mode that decides on the inference strategies. 

MACE (Models and Computer Examples) is a model builder that searches for finite models of first-order statements. It is based on a SAT solver, which implements the Davis, Putnam, Logemann and Loveland (DPLL) algorithm. The DPLL algorithm decides the satisfiability of a first-order propositional formula in conjunctive normal form (CNF) and builds a model for it if there exists a satisfying assignment \cite{nieuwenhuis2005abstract}. Basic operations of DPLL include unit propagation and pure literal elimination. A unit clause is a clause that contains only one single unassigned literal, say $l$. Once a unit clause is found, unit propagation replaces all occurrences of $l$s to true and $\neg l$s to false and then simplifies the resulting formula. A positive literal $l$ is pure if it only appears in one clause of the formula. Pure literal elimination assigns $l$ as true and removes the clause that contains it. Initially, the value for each literal in the given input is unassigned. DPLL first runs unit propagation and pure literal elimination while updating the current truth assignment. After simplifying the formula that results from the first step, it chooses an undefined literal and assigns a truth value to it. Next, DPLL calls itself recursively with the current propositional formula and truth assignment. The entire procedure repeats until all clauses are true under the current truth assignment, which is the satisfying assignment. Otherwise, it backtracks to a previous branching point and maps the literal to the opposite truth value. A first-order propositional formula is unsatisfiable if the DPLL procedure terminates without finding a satisfying assignment. 

MACE conducts model building in a five-step pipeline. In the first step MACE translates the given input, a first-order formula, to a set of first-order clauses using a subroutine of Otter. In the second step, MACE transforms the first-order clauses to a set of flattened, relational clauses. A flattened relational clause is similar to a first-order clause except that it does not contain any constant or function symbols. Essentially, each $n$-ary function symbol in the first-order clauses is replaced by an $n$+1-ary predicate symbol. For example, the function literal $g(x, y)=z$ is substituted with a three-place predicate $G(x,y,z)$ where $x$, $y$, and $z$ are variables. A constant symbol, say $e$, is rewritten to the predicate $E(z)$ where $z$ is a variable. In another example, a positive equality $\alpha$ = $\beta$ where $\alpha$ and $\beta$ are nonvariable is replaced by two clauses: $\alpha \neq x \vee \beta = x$ and  $\alpha = x \vee \beta \neq x$ where $x$ is a variable. In the third step, MACE generates a set of propositional clauses based on the flattened relational clauses and a given domain size $n$. Basically, MACE constructs all ground instances of the flattened relational clauses over the given domain, and then encodes each atom into a unique integer that becomes a propositional variable. In addition, MACE adds two constraints to each predicate introduced from the previous step. For instance, given the function literal $f(x,y) = z$ and its corresponding flattened relational clause $F(x,y,z)$, the first condition is that the clause $\neg F(x,y,z_{1}) \vee \neg F(x,y,z_{2})$ holds for any $z_{1}$ and $z_{2}$ where $z_{1} < z_{2}$. This guarantees that the last argument is a function of the first and second arguments. The second condition is that the clause $F(x,y,0) \vee F(x,y,1) \vee \cdots \vee F(x,y,n-1)$ holds. This ensures that the image of each function should be within the given domain. In the fourth step, MACE calls the DPLL procedure to search the model for the propositional clauses. Once a propositional model is found, the fifth step translates it back to the corresponding first-order model based on the encodings in the second and the third steps.

The PENG reasoner runs Otter and MACE in parallel in the process of consistency and informativity checking. Given a theory $\Psi$, the PENG reasoner uses Otter to detect its inconsistency while concurrently running MACE to detect its consistency. If Otter succeeds in finding a proof for $\neg \Psi$, then $\Psi$ is inconsistent. If MACE succeeds in constructing a model for $\Psi$, then $\Psi$ is consistent. Similarly, given a sentence $\psi$ and its previous context $\Phi$, if Otter succeeds in finding a proof for $\Phi \rightarrow \psi$, then $\psi$ is not informative. If MACE succeeds in constructing a model for $\Phi$ $\&$ $\neg \psi$, then $\psi$ is informative.

\subsection{Nonmonotonic Extensions of PENG}
The latest version of PENG incorporates typical nonmonotonicity in natural language and uses the answer set programming paradigm \cite{gelfond2012knowledge} to conduct nonmonotonic reasoning.
The syntax and semantics of answer set programming are described in detail in Appendix A. In this subsection, we will show some distinguishing features of ASP that can model the nonmonotonicity that occurs in natural language. 

\noindent
\textbf{Negation.} An ASP program can have two types of negations:  \textit{negation\ as\ failure} and strong negation. \textit{Negation\ as\ failure} is used to represent a sentence of the form ``there is no evidence that $\ldots$'' while a strong negation is used to represent a negation in natural language. For example, given the sentence, ``If Mary has an assignment to do and there is no evidence that the library is not open, then Mary will study in the library,'' it is translated into the following ASP rule
\begin{equation*}
study(Mary,library)\leftarrow not\ \neg open(library), has(Mary, assignment)
\end{equation*}
\noindent
\textbf{Constraints.} A constraint is defined as an ASP rule with an empty head. It is used to represent the sentences that denote restrictions in natural language. For instance, given the sentence, ``a person cannot be both a male and a female,'' it is translated into the following constraint:
\begin{equation*}
\leftarrow person(X), male(X), female(X).
\end{equation*}
\noindent
The rule enforces the \textit{answer\ set} to eliminate the atoms $person(s_{1})$, $male(s_{2})$, $female(s_{3})$ where $s_{1}=s_{2}$ and $s_{2}=s_{3}$.

\noindent
\textbf{Choice Rules.}
A choice rule is used to specify the number of head literals that may be included in an \textit{answer\ set} given that the body is true. For example, given the following sentences, 
\begin{enumerate}
\item There are two vertices: v1, v2 and v3.
\item There are three colors: red, green, and yellow.
\item Each vertex can be assigned with exactly one color.
\end{enumerate}
\noindent
they are translated into the following ASP rules,
\begin{enumerate}[label={},nolistsep]
\item $vertex(v1).\ vertex(v2).\ vertex(v3).$
\item $color(red).\ color(green).\ color(blue).$
\item $1\{assign(X,Y):color(Y)\}1\leftarrow vertex(X)$
\end{enumerate}
\noindent
where the third rule denotes a choice rule. It says that if $vertex(X)$ is in the \textit{answer\ set}, then for all $Y$'s such that $color(Y)$ is true the \textit{answer\ set} will choose to include one and only one head literal  $assign(X,Y)$. 

\noindent
\textbf{Defaults and Exceptions.}  The general form of a \textit{default} is represented as 
\begin{equation*}
p(X)\leftarrow c(X), not\ ab(d(X)), not\ \neg p(X).
\end{equation*}
\noindent
where $d$ stands for default, $ab$ represents abnormality, $not$ refers to \textit{negation\ as\ failure}, and $\neg$ denotes strong negation. The formula says that $X$ has the property $P$ if 1) $X$ is a class of $C$, 2) $X$ is not abnormal with respect to $d$, and 3) it cannot be shown that $X$ does not have property $P$. Given the \textit{default}, if $e(X)$ is a \textit{strong\ exception}, then it is represented as
\begin{equation*}
\neg p(X)\leftarrow e(X).
\end{equation*}
\noindent
If $e(X)$ is a \textit{weak\ exception}, it is represented as
\begin{equation*}
ab(d(X))\leftarrow not\ \neg e(X)
\end{equation*}
\noindent
The formula says $X$ is abnormal with respect to $d$ if there is no evidence that $X$ is not an \textit{exception}. 

Next, we will give three examples from \cite{schwitter2013working,schwitter2012answer,schwitter2013jobs} that show how ASP is used to represent knowledge with \textit{defaults}, \textit{exceptions}, \textit{choice\ rules}, and \textit{constraints}. 

\subsubsection{Defaults and Exceptions Use Case}
\begin{figure}[h]
    \centering
    \includegraphics[scale=1]{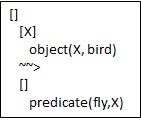}
    \caption{The DRS for the statement, ``Generally, birds fly.''}
    \label{peng-2}
\end{figure}

PENG extension in \cite{schwitter2013working} defines a \textit{default} as a statement that contains words such as \textit{generally}, \textit{normally}, or \textit{typically}. As compared to a strict rule, such as ``All birds fly,'' a default expresses a fact that is true in most cases but not always. For example, the statement, ``Generally, birds fly,'' indicates that most birds fly with a few exceptions. PENG allows two types of exceptions: strong exception and weak exception. A strong exception refutes the default conclusion and derives the opposite one, e.g., ``Penguins are birds. They do not fly.'' A weak exception makes the \textit{default} inapplicable without defeating the default conclusion. For instance, given the weak exception, ``An eagle is a bird. It is injured,'' it becomes unknown whether the eagle flies or not. To represent a \textit{default} in DRS, PENG introduces a new operator $\sim \sim >$. An example is given in Figure \ref{peng-2}, which shows the DRS for the sentence ``Generally, birds fly.''

Next, we will give an example from \cite{schwitter2013working} that shows how ASP is used to represent the semantics of PENG with \textit{defaults} and \textit{exceptions}. The knowledge base contains the following CNL sentences:
\begin{enumerate}
  \item Sam is a child.
  \item John is the father of Sam and Alice is the mother of Sam.
  \item Every father of a child is a parent of the child.
  \item Every mother of a child is a parent of the child.
  \item Parents of a child normally care about the child.
  \item John does not care about Sam.
  \item Alice is absent.
  \item If there is no evidence that a parent of a child is not absent then the parent abnormally cares about the child.
\end{enumerate}
\noindent
Sentence (1)-(4) denote strict rules, which can be represented in ASP as follows:
\begin{enumerate}[label={},nolistsep]
\item $child(sam)$
\item $father(john,sam)$
\item $mother(alice,sam)$
\item $parent(X,Y)\leftarrow father(X,Y),child(Y)$
\item $parent(X,Y)\leftarrow mother(X,Y),child(Y)$
\end{enumerate}
\noindent
Sentence (5) indicates a \textit{default}. It is translated into the following rule:
\begin{equation*}
care(X,Y)\leftarrow parent(X,Y), child(Y), not\ ab(d_{care}(X,Y)), not\ \neg care(X,Y)
\end{equation*}
\noindent
The formula says that $X$ cares about $Y$ if \romannumeral1) $X$ is a parent of $Y$, \romannumeral2) $Y$ is a child, \romannumeral3) there is no evidence that the pair  $<X,Y>$ is abnormal with respect to $d_{care}$, and \romannumeral4) there is no evidence that $X$ does not care about $Y$. Sentence (6) is a \textit{strong\ exception} to the \textit{default}. It is represented as a negative atom, $\neg care(john,sam)$. Sentence (8) denotes that sentence (7) is a \textit{weak\ exception} to the \textit{default}. They are represented as the following rules:
\begin{enumerate}[label={},nolistsep]
\item $absent(alice)$
\item $ab(d_{care}(X,Y))\leftarrow parent(X,Y), child(Y), not\ \neg absent(X)$
\end{enumerate}
\noindent
The second rule says that the pair $<X,Y>$ is abnormal with respect to $d_{care}$ if \romannumeral1) $X$ is a parent of $Y$, \romannumeral2) $Y$ is a child, \romannumeral3) there is no evidence that $X$ is not absent.

The \textit{answer\ set} for the above rules is 
\begin{enumerate}[label={},nolistsep]
\item $\{\ child(sam), father(john,sam), mother(alice,sam), absent(alice),$
\item $\ \ \ \neg care(john,sam), parent(john,sam), parent(alice,sam),$ 
\item $\ \ \ ab(d(care(alice,sam)))$, $ab(d(care(john,sam)))\}$
\end{enumerate}
\noindent
Given that $\neg care(john,sam)$ is in the \textit{answer\ set}, we can conclude that John does not care about Sam. Since neither $care(alice,sam)$ nor $\neg care(alice,sam)$ is in the \textit{answer\ set}, it is not known whether Alice cares about Sam or not.

\subsubsection{The Marathon Puzzle Use Case}
In \cite{schwitter2012answer}, Schwitter rewrites the marathon puzzle in PENG sentences which can be automatically and unambiguously translated into an ASP program to solve the problem. The marathon puzzle is described in natural language as 
\begin{enumerate}[label={},nolistsep]
\item Dominique, Ignace, Naren, Olivier, Philippe, and Pascal have arrived as the first six at the Paris marathon.
\item Reconstruct their arrival order from the following information:
\begin{enumerate}
\item Olivier has not arrived last.
\item Dominique, Pascal and Ignace have arrived before Naren and Olivier.
\item Dominique who was third last year has improved this year.
\item Philippe is among the first four.
\item Ignace has arrived neither in second nor third position.
\item Pascal has beaten Naren by three positions.
\item Neither Ignace nor Dominique are on the fourth position.
\end{enumerate}
\end{enumerate}
The puzzle is represented in PENG as follows:
\begin{enumerate}
\item Dominique, Ignace, Naren, Olivier, Philippe, and Pascal are runners.
\item There exist exactly six positions.
\item Every runner is allocated to exactly one position.
\item Reject that a runner R1 is allocated to a position and that another runner R2 is allocated to the same position and that R1 is not equal to R2.
\item Reject that Olivier is allocated to the sixth position.
\item If a runner R1 is allocated to a position P1 and another runner R2 is allocated to a position P2 and P1 is smaller than P2 then R1 is before R2.
\item Reject that Naren is before Dominique, Pascal, and Ignace.
\item Reject that Olivier is before Dominique, Pascal, and Ignace.
\item Reject that Dominique is allocated to a position that is greater than or equal to 3.
\item Reject that Philippe is allocated to a position that is greater than 4.
\item Reject that Ignace is allocated to the second position.
\item Reject that Ignace is allocated to the third position.
\item Reject that Pascal is allocated to a position P1 and that Naren is allocated to a position P2 and that P1 is not equal to P2 minus 3.
\item Reject that Ignace is allocated to the fourth position.
\item Reject that Dominique is allocated to the fourth position.
\end{enumerate}
Sentence (1) is translated into six ASP facts: $runner(dominique)$, $runner(ignace)$, $runner(naren)$, $runner(olivier)$, $runner(philippe)$, $runner(pascal)$. Sentence (2) is represented as an ASP fact with the range notation: $position(1\ldots 6)$. Sentence (3) denotes a choice rule with the cardinality constraint:
\begin{equation*}
1\{allocated\_to(X,Y):position(Y)\}1\leftarrow runner(X)
\end{equation*}
\noindent
The rule says that if $runner(X)$ is in the \textit{answer\ set}, then for al $Y$'s such that $position(Y)$ holds the \textit{answer\ set} will include exactly one head literal $allocated\_to(X,Y)$. Sentence (5) denotes a constraint. It is represented as 
\begin{equation*}
\leftarrow allocated\_to(olivier,6).
\end{equation*}
The constraint forces the atom $allocated\_to(olivier,6)$ to be excluded from the \textit{answer\ set}. Other constraints include sentence (7)-(15). They are represented in the similar way as sentence (5) does. Sentence (6) is a conditional sentence. It is represented as the following ASP rule:
\begin{equation*}
\begin{split}
before(X1,X2) \leftarrow runner(X1), position(Y1), allocated to(X1,Y1), \\runner(X2),  position(Y2), allocated to(X2,Y2), Y1 < Y2.
\end{split}
\end{equation*}
The complete ASP program is shown in Appendix B. There is one \textit{answer\ set} for the above ASP program where the position information  is $\{position(ignace,1)$, $position(dominique,2)$, $position(pascal,3)$, $position(philippe,4)$, $position(olivier,5)$, $position(naren,6)\}$.

\subsubsection{The Jobs Puzzle Use Case}
Just as the marathon puzzle, Schwitter also solves the jobs puzzle using ASP \cite{schwitter2013jobs}. The jobs puzzle is described as 
\begin{enumerate}
\item There are four people: Roberta, Thelma, Steve, and Pete.
\item Among them, they hold eight different jobs.
\item Each holds exactly two jobs.
\item The jobs are: chef, guard, nurse, telephone operator, police officer (gender not implied), teacher, actor, and boxer.
\item The job of nurse is held by a male.
\item The husband of the chef is the telephone operator.
\item Roberta is not a boxer.
\item Pete has no education past the ninth grade.
\item Roberta, the chef, and the police officer went golfing together.
\end{enumerate}
The corresponding CNL sentences of the jobs puzzle are shown below: 
\begin{enumerate}
\item Roberta is a person. Thelma is a person. Steve is a person. Pete is a person.
\item Roberta is female. Thelma is female.
\item Steve is male. Pete is male.
\item Exclude that a person is male and that the person is female.
\item If there is a job then exactly one person holds the job.
\item If there is a person then the person holds exactly two jobs.
\item Chef is a job. Guard is a job. Nurse is a job. Telephone operator is a job. Police officer is a job. Teacher is a job. Actor is a job. Boxer is a job.
\item If a person holds a job as nurse then the person is male.
\item If a person holds a job as actor then the person is male.
\item If a first person holds a job as chef and a second person holds a job as telephone operator then the second person is a husband of the first person.
\item If a first person is a husband of a second person then the first person is male.
\item If a first person is a husband of a second person then the second person is female.
\item Exclude that Roberta holds a job as boxer.
\item Exclude that Pete is educated.
\item If a person holds a job as nurse then the person is educated.
\item If a person holds a job as police officer then the person is educated.
\item If a person holds a job as teacher then the person is educated.
\item Exclude that Roberta holds a job as chef.
\item Exclude that Roberta holds a job as police officer.
\item Exclude that a person holds a job as chef and that the person holds a job as police officer.
\end{enumerate}
\noindent
For the above CNL sentences, sentence (1)-(3) and (7) denote ASP facts. Sentence (5)-(6) indicate choice rules. Sentence (4), (13)-(14), (18)-(20) signify constraints. Sentence (8)-(12), (15)-(17) represent conditional sentences. The complete ASP program is shown in Appendix B. There is one \textit{answer\ set} for the above problem. The job information for the \textit{answer\ set} is $\{holds(thelma,chef)$, $holds(roberta,guard)$, $holds(steve,nurse)$, $holds(pete,operator)$, $holds(steve,police)$, $holds(roberta,teacher)$, $holds(pete,actor)$, $holds(thelma,boxer)\}$ where predicate $holds(X,Y)$ indicates that $X$ holds the job $Y$.

\section{Computer-Processable Language}

In this section, we discuss Computer-Processable Language (CPL), which was developed by Peter Clark at University of Texas \cite{clark2005acquiring}. The first subsection introduces CPL's language properties, the second subsection discusses its semantic interpretations, and the third gives an overview of CPL's inference system.
\subsection{Language Properties}
The vocabulary of CPL is based on a pre-defined Component Library (CLib) ontology, which was developed at UT Austin \cite{barker2001library}. Unlike ACE or PENG, CPL does not allow users to extend the vocabulary. Instead, it uses WordNet to map the words that are outside of the target ontology to the closest concepts within the CLib ontology. In particular, modal words, such as probably and mostly, are not in CPL because they cannot be represented in first-order logic. 

CPL accepts three types of sentences: ground facts, rules, and questions. Ground facts are basic CPL sentences that have the following forms:
\begin{quote}
\textbf{There is $\mid$ are} \textit{NP}\\
\textit{NP verb [NP]} $\textit{[PP]}^{\star}$\\
\textit{NP} \textbf{is $\mid$ are} \textit{passive-verb} \textit{[\textbf{by} NP]} $\textit{[PP]}^{\star}$
\end{quote}
\noindent
where \textit{NP} denotes a noun phrase, \textit{PP} refers to a prepositional phrase, \textit{[NP]} signifies that the noun phrase is optional, and $\textit{[PP]}^{\star}$ indicates that there can be zero or more prepositional phrases. Rules are of the form:

\textbf{IF} \textit{Sentence} \textit{[}\textbf{AND} $\textit{Sentence]}^{\star}$ \textbf{THEN} \textit{Sentence} \textit{[}\textbf{AND} $\textit{Sentence]}^{\star}$

\noindent
where \textit{Sentence} denotes a basic CPL sentence, and \textit{[}\textbf{AND} $\textit{Sentence]}^{\star}$ denotes zero or multiple basic CPL sentences. CPL accepts five forms of questions that begin with ``What is,'' ``What are,'' ``How many,'' ``How much,'' or ``Is it true.'' 

For ground facts, objects are considered existentially quantified. To express universally quantified objects, users must use $\textbf{IF}\ \ldots\ \textbf{THEN}\ \ldots\ $ rules. Words that indicate universal quantifiers (e.g. all, every, most) are banned in CPL. Compared to ACE and PENG, which use the words such as ``all'' and ``every'' to describe universally quantified objects, CPL makes the language more redundant and stilted. For example, given the ACE/PENG sentence, ``every student that gets an A is smart,'' it is written in CPL as ``\textbf{IF} a student gets an A \textbf{THEN} the student is smart.''  

Unlike ACE or PENG, CPL does not allow pronouns. Users must use either definite reference or ordinal reference to indicate previously mentioned objects. A definite reference is resolved by searching for the most recent, previously mentioned noun. For example, given the sentences, ''There is a dog in the park. The dog is cute,'' ``the dog'' in the second sentence refers to the dog mentioned in the first sentence. An ordinal reference is resolved in a similar way, except that it counts the number of occurrences of the mentioned noun from the beginning of a paragraph and chooses the appropriate occurrence as defined by the ordinal. For example, given the sentences, ``Tom has a dog. Mary has a dog. The first dog is cute. The second dog is smart,'' to resolve ``the first dog'' in the third sentence, the CPL interpreter starts from the first sentence and finds the first occurrence of dog, Tom's dog. Similarly, CPL resolves ``the second dog'' in the fourth sentence by searching the second occurrence of dog from the beginning of the paragraph. 

\subsection{Semantic Interpretations}
The semantics of CPL are represented by KM (Knowledge Machine) sentences. KM is a powerful frame-based knowledge representation language \cite{clark2004km}. It represents first-order logic clauses in LISP-like syntax. 

The CPL interpreter translates a CPL sentence into KM sentences in three steps. First, the interpreter uses a bottom-up, broad coverage chart parser, called SAPIR, to parse a CPL sentence and then generates a logical form (LF) \cite{harrison1986new}. An LF is a simplified and normalized tree structure with logic-type elements \cite{clark2008boeing}. It uses variables, which are prefixed by underscores, to represent noun phrases. An LF captures the syntactic properties and relations of the sentence, including tense, aspect, polarity, and a tag set consisting of S (sentence), PP (prepositional phrase), NN (noun compound), PN (proper name), PLUR (plural), and PLUR-N (numbered plural). For example, the sentence ``the cat sat on the reed mat'' is shown below in the LF form:

\begin{quote}
((VAR \texttt{\_X1} ``the'' ``cat'')\\
(VAR \texttt{\_X2} ``the'' ``mat'' (NN ``reed'' ``mat''))\\
(S (PAST) \texttt{\_X1} ``sit'' (PP  ``on'' \texttt{\_X2})))
\end{quote}

\noindent
The first line says \texttt{\_X1} is a variable that represents a noun phrase, ``the cat.'' The second line says \romannumeral1) \texttt{\_X2} is a variable that denotes the noun phrase, ``the reed mat,'' \romannumeral2) ``reed mat'' is a noun compound where ``reed'' is the noun modifier of ``mat.'' The third line says \romannumeral1) the tense of the sentence is simple past, \romannumeral2) \texttt{\_X1}, which denotes ``the cat,'' is the subject of the \textit{sitting} event,  \romannumeral3) the prepositional phrase, ``on the reed mat,'' is the prepositional modifier of ``sit.''

Second, an initial logic generator is used to transform the LF into ground logical assertions (KM sentences) by applying a set of simple, syntactic rewrite rules. Logical assertions are binary predicates that represent the syntactic relations and the prepositions of the LF generated in Step 1, including subject (syntactic subject), sobject (syntactic object), mod (modifier), all prepositions, etc.. Other information, such as part of speech, tense, polarity, and aspect, is omitted. For the above example, ``the cat sat on the reed mat,'' its logical assertions are shown below:
\begin{quote}
subject(\texttt{\_Sit1}, \texttt{\_Cat1})\\
``on''(\texttt{\_Sit1}, \texttt{\_Mat1})\\
mod(\texttt{\_Mat1}, \texttt{\_Reed1})
\end{quote}
\noindent
Here, \texttt{\_Sit1}, \texttt{\_Cat1}, \texttt{\_Mat1}, \texttt{\_Reed1} refer to the words ``sit,'' ``cat,'' ``mat,'' and ``reed'' respectively in the LF. They are \textit{Skolem} constants that denote the instances of some concepts (classes) in the target ontology. These concepts are determined in the third step. Predicates \textit{subject} and \textit{mod} denote the syntactic relations of the sentence. Predicate ``on'' signifies the preposition ``on'' in the LF. 

Third, subsequent post-processing is performed based on the logical assertions generated in Step 2, including word sense disambiguation, semantic role labelling, and structural re-organization. For word sense disambiguation, each  \textit{Skolem} instance is assigned with a concept in the target ontology, based on the word each \textit{Skolem} instance corresponds to and its part of speech information. Essentially, the interpreter selects the most common synset (a group of synonymous words) for a word in WordNet and then maps the WordNet synset to the corresponding concept in CLib ontology. For the above example, four additional sentences are added to the knowledge base:
\begin{quote}
isa(\texttt{\_Sit1}, \texttt{Sit\_n1})\\
isa(\texttt{\_Cat1}, \texttt{Cat\_n1})\\
isa(\texttt{\_Mat1}, \texttt{Mat\_n1})\\
isa(\texttt{\_Reed1}, \texttt{Reed\_n1})
\end{quote}
\noindent
where \texttt{Sit\_n1}, \texttt{Cat\_n1}, \texttt{Mat\_n1}, \texttt{Reed\_n1} denote the CLib ontology concepts. The predicate \textit{isa} is a binary relation. It indicates that an entity is an instance of a class.

By using word sense disambiguation, CPL is capable of identifying synonyms. Different words that denote the same concept will be mapped to the same concept in CLib ontology, and thus will be co-referenced. This is more advantageous than ACE and PENG. For instance, ACE cannot recognize synonyms. Hence, words that represent the same concept will be regarded as different if they are distinct. PENG has a list of pre-defined synonyms in its dictionary, but its synonym information is much less than that of WordNet, which contains 155,287 words organized in 117,659 synsets for a total of 206,941 word-sense pairs \cite{miller1995wordnet}. 

Next, semantic role labelling (SRL) is performed to replace some syntactic relations with semantically meaningful relations. For the above example, the binary predicates generated in Step 2 are replaced by the following:
\begin{quote}
agent(\texttt{\_Sit1}, \texttt{\_Cat1})\\
location(\texttt{\_Sit1}, \texttt{\_Mat1})\\
material(\texttt{\_Mat1}, \texttt{\_Reed1})
\end{quote}
\noindent
where \textit{agent} indicates that an entity initiates, performs or causes an event, \textit{location} signifies that an event ends at a place, and \textit{material} shows that an entity is made of another entity. 

Finally, structural re-organization is deployed. For example, given two binary predicates subject(\texttt{\_Be1}, \texttt{\_Rose2}) and object(\texttt{\_Be1}, \texttt{\_Red3}), structural re-organization will merge them into a single predicate, ``be''(\texttt{\_Rose2}, \texttt{\_Red3}). Another example is that any \textit{equal} predicate, say equal(\texttt{\_Color1}, \texttt{\_Red2}), will be removed from the logical assertions and the occurrences of \texttt{\_Color1} will be replaced by \texttt{\_Red2} as well. 

Similar to ACE and PENG, CPL will display the final interpretations and the paraphrases of the input to users for judgement. If the users do not accept the interpretation, they can rephrase the original sentence and/or modify the word senses and semantic relations. 

\subsection{Reasoning Services}
For CPL's reasoning service, users can compose questions in CPL using one of the five forms described in section 4.1. Compared to ACE and PENG, CPL does not support consistency checking, informativity checking, nor does it support theorem proving for a given set of sentences. 

CPL deploys KM as the underlying inference system. As is mentioned in section 4.2, KM is a frame-based knowledge representation language implemented in Common Lisp. The basic representation unit is a \textit{frame}, which consists of a set of \textit{slots} and \textit{values}. \textit{Slots} denote binary relations between \textit{instances}. They can represent the properties of a \textit{class} or an \textit{instance}. KM defines three types of properties for a \textit{class}: its own properties, assertional properties, and definitional properties. A \textit{class}'s own properties describe the meta-data of the \textit{class} itself. They do not apply to any member of the \textit{class}. \textit{Assertional\ properties} denote the properties that are implied by the membership of the \textit{class}. They are formulated as uni-directional implications:

 $\forall \ x\ (C(x) \rightarrow s_{1}(x,v_{11}), \ldots, s_{1}(x,v_{1k_{1}}), \ \ldots ,\ s_{n}(x,v_{n1}), \ldots, s_{n}(x,v_{nk_{n}}))$

\noindent
where $C$ denotes a \textit{class}, $s_{i}\ (i = 1,2,\ldots,n)$ represents a \textit{slot} indicating one of the \textit{class}'s assertional properties, $v_{ij} (i = 1,2,\ldots,n, j = 1,2,\ldots,k_{i})$ is a value in \textit{slot} $s_{i}$, and $s_{i}(x,v_{ij})$ is a binary relation that holds between \textit{instances} of $C$ and the value of \textit{slot} $s_{i}$. \textit{Definitional\ properties} are the properties that are both implied by the membership of the \textit{class} and are sufficient to conclude the \textit{class} membership of an \textit{instance}. They are formulated as bi-directional implications:

 $\forall \ x\ (C(x) \leftrightarrow s_{1}(x,v_{11}), \ldots, s_{1}(x,v_{1k_{1}}), \ \ldots ,\ s_{n}(x,v_{n1}), \ldots, s_{n}(x,v_{nk_{n}}))$

\noindent
where $s_{i}\ (i = 1,2,\ldots,n)$ represents a \textit{slot} indicating one of the \textit{class}'s definitional properties. An \textit{instance} can be classified into \textit{class} $C$ if it satisfies all of $C$'s definitional properties. Both assertional and definitional properties apply to every member of the \textit{class}. \textit{Classes} can inherit the properties from their \textit{superclasses}. Once an \textit{instance} is created, it will inherit the properties from all of its \textit{superclasses}. 

KM has a mechanism, called \textit{automatic\ classification}, which will automatically attempt to re-classify an \textit{instance} into the most specific \textit{class} once it is created or modified. For example, given the \textit{classes} Square and Rectangle where Square is defined to be the \textit{subclass} of Rectangle with the definitional property of equal length and width, if the user creates an \textit{instance} of Rectangle with its length equal to its width, KM will search the inheritance hierarchy and re-classify the \textit{instance} into the Square \textit{class}. Later, if the user modifies the \textit{instance}, say changing its length to be smaller than its width, KM will again search the inheritance hierarchy and re-classify the \textit{instance} into the Rectangle \textit{class}.

KM performs inference when a query is issued to the knowledge base. A query is of the form: 
 
$(the\ slot_{n}\ of \ (\ \ldots\  (the\ slot_{2}\ of \ (the\ slot_{1}\ of\ expr ))))$

\noindent
where $slot_{i}\ (i = 1,2,\ldots,n)$ represents a \textit{slot} and $expr$ denotes a KM expression that evaluates to one or multiple \textit{instances}. The semantics of a query are formulated as an access path. It is a join of binary predicates of the form:

$P_{1}(X_{0},x_{1}),P_{2}(x_{1},x_{2}),\ldots,P_{n}(x_{n-1},x_{n})$

\noindent
where $X_{0}$ is a constant, and $x_{i}$ $(i = 1,2,\ldots,n)$ is a free variable. It denotes a set S of values for $x_{n}$ such that 

$\forall \ x_{n}\ (x_{n} \in S \leftrightarrow \exists x_{1},\ldots,x_{n-1}\ P_{1}(X_{0},x_{1}),P_{2}(x_{1},x_{2}),\ldots,P_{n}(x_{n-1},x_{n}))$

KM evaluates a query from right to left in $n$ iterations. In iteration 1, KM processes $(the\ slot_{1}\ of\ expr )$ by computing the value of $expr$ and then finding the value of $slot_{1}$ for each \textit{instance} returned by $expr$. The automatic classification mechanism is always implicitly applied to each \textit{instance} generated in the whole process. The results are fed as input to iteration 2. In iteration 2, KM finds the value of $slot_{2}$ for each \textit{instance} in the input. The results are again passed to iteration 3. This process repeats until it finishes processing $slot_{n}$.

\section{Nonmonotonicity in Controlled Natural Languages}
According to \cite{sep-logic-nonmonotonic}, nonmonotonic logics are a family of logics that are designed to represent the kind of defeasible inference in everyday life, where reasoners draw a set of conclusions that are justified by the given knowledge base. These conclusions may be invalidated with the addition of more knowledge. Examples of nonmonotonic logic frameworks include circumscription\cite{mccarthy1980circumscription}, default logic \cite{reiter1980logic}, autoepistemic logic \cite{mcdermott1982nonmonotonic}, etc.. In this section, we first study two types of nonmonotonic phenomena that occur in natural languages: \textit{defaults\ and\ exceptions} and \textit{conversational\ implicatures}. Then propose their adapted representations in CNLs and the corresponding formalizations in nonmonotonic reasoning frameworks.

\subsection{Defaults and Exceptions}
A default is a statement that is true by default about a collection of instances but may be defeated by information about some specific instances. The latter are called exceptions. This contrasts with a definite statement in the real world where the occurrences of such instances will falsify the validity of the statement. In natural languages, a default can be empirically categorized into two types: one that is directly declared and one that is indirectly specified. A default is directly declared if it is a statement that generalizes what an object does and contains particular words (e.g. generally, typically, normally) \cite{schwitter2013working}. For instance, ``\textit{Normally}, birds fly,'' ``If she has an essay to write, \textit{typically} she will study in the library,'' etc. A default is  stated indirectly if it does not contain any keywords but can be identified by the context or based on some commonsense knowledge, e.g., ``There is a lot of rain in Seattle.'' A default hides a number of unstated assumptions; in the previous case, e.g., ``There is no drought in Seattle,'' ``The climate in Seattle does not change,'' etc.. The default is defeated if one of the assumptions does not hold. Similarly, exceptions can be  noted directly with the keyword, such as \textit{except}, or  expressed indirectly. Examples are shown below,

\begin{enumerate}
\item If she has an essay to write, she will study late in the library except on weekends.
\end{enumerate} 

\begin{enumerate}[label={2.\alph*},nolistsep]
\item Typically, birds fly.
\item Penguins do not fly. 
\end{enumerate}

\begin{enumerate}[label={3.\alph*},nolistsep]
\item If Mary has a writing assignment, typically she will study in the library.
\item If Mary has a coding assignment, normally she will study in CS lab. 
\end{enumerate}

\noindent
Sentence (1) contains a default statement and its exception information is directly specified by the keyword ``except.'' The sentence says that normally she will study in the library. But if it is a weekend, she will not study in the library. Sentence (2.a) denotes a default, which generalizes what birds do. For sentence (2.b), although it does not contain any keywords to indicate an exception to (2.a), we can still identify the exception  based on the context and on commensense knowledge. Sentence (3.a) and (3.b) are ambiguous. The default conclusions drawn from (3.a) and (3.b) are incompatible with each other because Mary cannot study in two different locations at the same time. However, there is no information of what is an exception to what. Hence, it is impossible to decide where Mary will study when she has both a writing and a coding assignment to do.

We adapt defaults and exceptions that occur in natural languages to the design of CNLs. Unlike natural languages, CNLs have to be precise and unambiguous without losing the naturalness of the language. To the best of our knowledge, PENG is the only CNL that incorporates defaults and supports nonmonotonic reasoning. As  described in Section 3.4, PENG defines a default as a general statement that contains words, such as \textit{normally,\ generally,} and \textit{typically}. There are two types of exceptions in PENG: strong exception and weak exception. A strong exception contradicts the conclusion generated by the default. A weak exception makes the default inapplicable. Defaults and exceptions are formalized in ASP programs. One problem with PENG is its stilted way to indicate that a statement is a weak exception to the default. An example is shown below: 

\begin{enumerate}[label={4.\alph*},nolistsep]
\item Parents of a child normally care about the child.
\item Tom is a parent of a child.
\item Tom does not care about his child. 
\item Alice is a parent of a child. 
\item Alice is absent.
\end{enumerate}
\noindent
Sentence 4.a is a default, which is formalized as 

care(X,Y) :- parent(X,Y), child(Y), not ab(d(care(X,Y))), not $\neg$care(X,Y).

\noindent
where $d$ stands for default, $ab$ represents abnormal, $not$ signifies \textit{negation\ as\ failure}, and $\neg$ refers to strong negation. Sentence (4.c) denotes a strong exception to (4.a). In order to ensure that sentence 4.e is a weak exception to 4.a, users have to write the \textit{cancellation\ rule} in CNL as ``If there is no evidence that a parent of a child is not absent then the parent abnormally cares about the child.'' This makes the language processor translate it into the following ASP rule:

ab(d(care(X,Y))) :- parent(X,Y), child(Y), not $\neg$absent(X).

\noindent
The way the cancellation rule is expressed in PENG is just a direct translation of the intended ASP rule, which is rather unintuitive.  

We propose a different approach to the representations and formalizations of \textit{defaults\ and\ exceptions} in CNLs. Each CNL sentence is associated with a unique identifier, which denotes the location of the sentence in the text. The identifier can be an English phrase, e.g. the second sentence of the third paragraph, or a  user-defined label, such as para3sent2, where \textit{para} stands for paragraph and \textit{sent} stands for sentence. We assume that CNL sentences are defeasible by default and therefore can be defeated. To denote a definite statement, users must add the keyword, \textbf{strict}, at the beginning of the sentence, e.g.,

\begin{enumerate}
\setcounter{enumi}{4}
\item (\textbf{strict}) Obama won the presidential election in 2012. 
\end{enumerate}

Exception information has to be  mentioned directly. A default can have three types of exceptions: refutation, rebuttal, and cancellation. The refutation of a default offers a conclusion that is incompatible with the conclusion drawn from the default and will override the default. Here, we assume that the ``refutation'' relation is transitive. That is, if statement A refutes B and B refutes C, statement A refutes C as well. The structure of a refutation is a CNL statement proceeded by its reference information, which specifies the corresponding sentences the current statement causes an exception to. The reference information is represented by either template 6.a or 6.b,
\begin{enumerate}[label={6.\alph*},nolistsep]
\item \textbf{\textit{except}}
\item \textbf{\textit{exception to}} $id_{1},\ id_{2}, \ldots ,\ id_{n}$
\end{enumerate} 
\noindent
where $id_{i}$'s ($i=1,2,\ldots,n$) denote sentence identifiers. Template 6.a indicates that the current sentence is  a refutation to its previous one. Template 6.b says that the sentence is a refutation to sentences with ids $id_{1},\ id_{2}, \ldots ,\ id_{n}$. The rebuttal of a default is also a conclusion that is incompatible with the default. However, there is no information on which conclusion has more weight. Hence, no conclusion can be drawn from the default and its rebuttal. To specify the situation where two conclusions are incompatible with each other, we write it as a  CNL statement preceded by the keyword, \textbf{conflict constraint}, e.g.,

\begin{enumerate}[label={7.\alph*},nolistsep]
\item Tom votes for Obama.
\item Tom votes for Romney.
\item Obama is a candidate.
\item Romney is a candidate.
\item (\textbf{conflict constraint}) A person can vote for \textbf{at most} one candidate.
\end{enumerate} 
\noindent
According to 7.e, sentence 7.a and 7.b are incompatible and therefore rebut each other. Thus, both 7.a and 7.b are false. The cancellation of a default simply renders the default inapplicable and therefore cancels the default conclusion. A cancellation is represented as a conditional of the form:

If \textit{P}, then \textbf{cancel} $id_{1},\ id_{2}, \ldots ,\ id_{n}$.

\noindent
where \textit{P} denotes the premises of the conditional. Note, all exceptions are defeasible. They can be defeated by other statements as well. In this case, an exception can be used to defeat other statements only when the exception itself is not defeated, e.g.,

\begin{enumerate}[label={8.\alph*},nolistsep]
\item If John gets an offer from BOA, John will join BOA.
\item (\textbf{except}) If John gets an offer from Amazon, John will join Amazon.
\item John gets an offer from BOA.
\item John gets an offer from Amazon.
\item Amazon goes bankcruptcy.
\item If Amazon goes bankcruptcy, then \textbf{cancel} 8.b.
\end{enumerate} 
\noindent
Sentence 8.b causes an exception to 8.a. However, based on 8.e and 8.f, sentence 8.b is defeated. In this case, sentence 8.b cannot be used to refute 8.a. Hence, we can conclude that ``John will join BOA.''

We formalize CNL sentences using \textit{Logic\ Programming\ with\ Defaults\ and\ Argumentation\ Theories} (LPDA) \cite{wan2009logic}, which is a powerful framework for defeasible reasoning based on well-founded semantics \cite{przymusinski1994well}. LPDA has well-defined semantics for defaults and exceptions. There are two types of rules: strict rules and defeasible rules. Each LPDA program is accompanied with an argumentation theory, which decides when a defeasible rule is defeated. There are three special predicates: \textbf{opposes}, \textbf{overrides}, and \textbf{cancel} where \textbf{opposes} indicates the literals that are incompatible with each other,  \textbf{overrides} denotes a binary relation between defeasible rules indicating priority, and \textbf{cancel} cancels a defeasible rule. We use predicate \textbf{overrides} and \textbf{cancel} to formalize the ``refutation'' and ``cancellation'' relations between a default and its exceptions respectively. We use predicate \textbf{opposes} to formalize the incompatibilities between a defaults and rebuttals.

Next, we give a complicated example that includes three types of exceptions and show how they are formalized in LPDA to perform defeasible reasoning.

\begin{enumerate}[label={9.\alph*},nolistsep]
\item John is a store member.
\item John is an SBU employee.
\item John buys a can of coke.
\item John buys a lobster.
\item Mary is a store member.
\item Mary is an SBU employee.
\item Mary buys  salmon.
\item Mary is in the blacklist.
\item A Coke is a beverage.
\item Lobster is seafood.
\item Salmon is seafood.
\item If a customer buys a beverage, the customer gets a discount of \$1.50.
\item (\textbf{except}) If a customer is a store member and buys a beverage, the customer gets a discount of \$2.50.
\item If a customer is a store member and buys seafood, the customer gets a discount of \$7.50.
\item If a customer is an SBU employee and buys seafood, the customer gets a discount of \$5.00.
\item If a store member is in the blacklist, then \textbf{cancel} 9.m, 9.n.
\item (\textbf{conflict\ constraint}) A customer gets at most one discount for any product.
\end{enumerate} 
 
\noindent
The CNL sentences are formalized in LPDA as follows:
\begin{enumerate}[label={},nolistsep]
\item member(John).
\item sbuemployee(John).
\item buy(John,coke).
\item buy(John,lobster).
\item member(Mary).
\item sbuemployee(Mary).
\item buy(Mary,salmon).
\item blacklist(Mary).
\item beverage(coke).
\item seafood(lobster).
\item seafood(salmon).
\item @\{r1\} discount(?Customer,?Product,?Amount):-
\item $\ \ \ \ \ \  \ \ \ \ \ \ \ \ \ \ \ \ \ \ $buy(?Customer,?Product),beverage(?Product),?Amount \textit{is} 1.50.
\item @\{r2\} discount(?Customer,?Product,?Amount):-
\item $\ \ \ \ \ \  \ \ \ \ \ \ \ \ \ \ \ \ \ \ $buy(?Customer,?Product),beverage(?Product),
\item $\ \ \ \ \ \  \ \ \ \ \ \ \ \ \ \ \ \ \ \ $member(?Customer),?Amount \textit{is} 2.50.
\item @\{r3\} discount(?Customer,?Product,?Amount):-
\item $\ \ \ \ \ \  \ \ \ \ \ \ \ \ \ \ \ \ \ \ $buy(?Customer,?Product),seafood(?Product),
\item $\ \ \ \ \ \  \ \ \ \ \ \ \ \ \ \ \ \ \ \ $member(?Customer),?Amount \textit{is} 7.50.
\item @\{r4\} discount(?Customer,?Product,?Amount):-
\item $\ \ \ \ \ \  \ \ \ \ \ \ \ \ \ \ \ \ \ \ $buy(?Customer,?Product),seafood(?Product),
\item $\ \ \ \ \ \  \ \ \ \ \ \ \ \ \ \ \ \ \ \ $subemployee(?Customer),?Amount \textit{is} 5.00.
\item cancel(r2):-member(?Customer),blacklist(?Customer).
\item cancel(r3):-member(?Customer),blacklist(?Customer).
\item opposes(discount(?Customer,?Product,?Amount1),
\item $\ \ \ \ \ \  \ \ \ \ \ \ \ \ \ \ \ \ \ \ $discount(?Customer,?Product,?Amount2)):-
\item $\ \ \ \ \ \  \ \ \ \ \ \ \ \ \ \ \ \ \ \ \ \ \ \ \ \  \ \ \ \ \ \ \ \ \ \ \ \ \ \ $buy(?Customer,?Product),?Amount1 != ?Amount2.
\item overrides(r2,r1).
\end{enumerate}
\noindent
For the above example, if the user asks ``How much discount does John get for buying a coke,'' it will answer \$2.50. Sentence 9.l and 9.m give rise to the conclusions that John gets a discount of \$2.50 and \$1.50 respectively for buying a coke. According to 9.q, they are incompatible because John cannot get two discounts for buying one product. However, since sentence 9.m refutes 9.l, the conclusion drawn from 9.m will override the one made from 9.l. Thus, we have the conclusion that John gets a discount of \$2.50 for buying a coke. Next, if the user asks ``How much discount does John get for buying a lobster,'' it will return no answer. Just like the previous case, sentence 9.n and 9.o generate incompatible conclusions: John gets a discount of \$7.50 and \$5.00 for buying a lobster. Without any refutation information between 9.n and 9.o, the conclusion that John gets a discount for buying a lobster cannot be justified. Finally,  if the user asks ``How much discount does Mary get for buying  salmon,'' it will answer \$5.00. Even though 9.n and 9.o generate incompatible conclusions for Mary's purchase, since Mary is in the blacklist, sentence 9.n will be defeated by 9.p. In this case, sentence 9.n cannot be used to rebut 9.o. Thus, we have the conclusion that Mary gets a discount of \$5.00 for buying  salmon.

\subsection{Conversational Implicatures}

The concept of \textit{conversational\ implicatures} was proposed and systematically studied by H. P. Grice \cite{grice20134}. Grice stated that \textit{conversational\ implicatures} are a component of the meaning of the utterances that are drawn not only from the literal meaning of the given linguistic expressions, but also based on some general principles of conversational rationality. These principles are also known as Grice's maxims:

\begin{enumerate}[label=\alph*]
  \item Quantity
  	\begin{enumerate}
    		\item Make your contribution as informative as is required.
    		\item Do not make your contribution more informative than is required.
  	\end{enumerate}
  \item Quality
	\begin{enumerate}
		\item Do not say what you believe to be false.
		\item Do not say that for which you lack adequate evidence.
	\end{enumerate}
  \item Relation
	\begin{enumerate}
		\item Be relevant.
	\end{enumerate}
   \item Manner
	\begin{enumerate}
		\item Avoid obscurity of expression.
		\item Avoid ambiguity.
		\item Be brief.
		\item Be orderly.
	\end{enumerate}
\end{enumerate}

\noindent
Speakers are presumed to observe and obey the above maxims in their communications. Based on this assumption, when a speaker appears to break one of the maxims, it will cause the interpreter to make some inferences regarding what the speaker really means. For instance,
\begin{enumerate}
\setcounter{enumi}{9}
\item The final exam will take place in the CS building or the main library.
\end{enumerate}

\begin{enumerate}
\setcounter{enumi}{10}
\item Speaker A: Students said the exam covered all chapters in the book and was very difficult. \\Speaker B: It covered all chapters in the book.
\end{enumerate}
\noindent
Sentence 10 says that the speaker thinks both places are possible. Hence, according to the maxim of quantity, we can conclude that the speaker does not know exactly where the exam takes place. In sentence 11, speaker A said that the exam not only covered all chapters but also was very difficult. However, speaker B only mentioned that the exam covered all chapters. Again, according to the maxim of quantity, we can conclude that the exam covered all chapters but was not necessarily difficult. Conversational implicatures are defeasible in nature. This contrasts with the semantics of entailment in classical logic. For instance, if speaker B adds that ``in fact, the exam was also very difficult,'' it will cancel the implicature that ``the exam was not difficult.'' 

Grice divides conversational implicatures into two classes: \textit{generalized\ implicatures} and \textit{particularized\ implicatures}. \textit{Generalized\ implicatures} do not strongly depend on the context. They arise only when particular lexical items are used in the sentence. \textit{Particularized\ implicatures}, on the other hand, are based on specific context and arise only when such particular context occurs. In the previous example, sentence 10 denotes a \textit{generalized\ implicature}, which is triggered by the connective \textit{or}. Sentence 11 denotes a \textit{particularized\ implicature} where the implicature generated from B's utterance arises from the context A provides. Generalized implicatures can be further divided into many sub-classes, e.g., \textit{scalar\ implicatures} and \textit{clausal\ implicatures}. For the rest of this subsection, we will focus on modelling scalar implicatures using nonmonotonic logics.

Scalar implicatures are a class of generalized implicatures derived from the maxim of quantity. They are based on implicature scales, each denoting a set of lexical items ordered by their informativeness, e.g., $<$a few, some, many, all$>$. Given a sentence that uses one of the lexical items in the implicature scale, its corresponding weaker (and stronger) sentences are defined as those where such lexical items are substituted for the weaker (and stronger) ones in the implicature scale. The theory of scalar implicatures says that \romannumeral1) all weaker sentences are entailed by the given sentence, and \romannumeral2) all stronger sentence are false by default. 
For instance, based on the implicature scale, $<$a few, some, many, all$>$, the sentence, ``Many students passed the exam,''  entails that \romannumeral1) ``Some students passed the exam,'' and \romannumeral2) ``A few students passed the exam.'' Besides, it also implies that ``Not all students passed the exam.''

One of the previous works that use nonmonotonic logics to model \textit{scalar\ implicatures} is Wainer's compositional approach, which is based on circumscription \cite{wainer2007modeling}. Wainer defines the \textit{extended\ meaning} of a sentence as a semantic content of the sentence itself combined with (one of) its \textit{generalized\ implicatures}. The general form of the \textit{extended\ meaning} is represented as 
\begin{enumerate}[label={},nolistsep]
\item $\alpha \wedge (\neg abn(c_{1})\rightarrow \hat{\beta})$
\end{enumerate}
\noindent
Here, $\alpha$ and $\hat{\beta}$ denote the semantic content of the sentence and (one of) its \textit{generalized\ implicatures} respectively. The predicate $abn$ is the \textit{abnormality\ predicate}, which is an \textit{ad\ hoc} symbol that does not appear in $\alpha$ or $\hat{\beta}$. Similarly, the constant $c_{1}$ is also an \textit{ad\ hoc} symbol that is only used in one specific formula. The \textit{extended\ meaning} of the sentence is obtained by circumscribing predicate $abn$, that is, by minimizing the extension of predicate $abn$. Details of circumscription can be found in Appendix A.

As before, we model \textit{scalar\ implicatures} using LPDA. For the implicature scales related to quantifiers, e.g., $<$a few, some, many, all$>$, our current approach only supports one instance, $<$some, all$>$, where \textit{some} and \textit{all} correspond to the existential and universal quantifiers respectively in first-order logic. For example, given the sentence, ``Some students pass exam1,'' its \textit{extended\ meaning} is ``Some students pass exam1. There exists at least one student who does not pass exam1.'' The \textit{extended\ meaning} is formalized in LPDA as shown below:

\begin{enumerate}[label={},nolistsep]
\item student(\#1).
\item pass(\#1,e1).
\item student(\#2).
\item @r1 neg pass(\#2,e1).
\end{enumerate}

\noindent
where rule $r1$ indicates a defeasible rule and \#1 and \#2 are \textit{Skolem} constants. The first two rules represent the semantic content of the original sentence. The third and fourth rules represent the semantic content of the \textit{scalar\ implicature}. If later, the speaker adds that ``in fact, all students pass the exam,'' the following rule will be added,

\begin{enumerate}[label={},nolistsep]
\item pass(?x,e1):-student(?x).
\end{enumerate}

\noindent
This will refute the default conclusion that ``there is a student who did not pass exam1.'' 

We can also model the class of implicature scales related to predicates, e.g., $<$cute, beautiful, stupendous$>$. For example, given the sentence, ``Mary is beautiful,'' its \textit{extended\ meaning} is ``Mary is beautiful. Mary is not stupendous.'' The semantics of the implicature scale together with the \textit{extended\ meaning} of the sentence is formalized as

\begin{enumerate}[label={},nolistsep]
\item cute(?x):-beautiful(?x).
\item beautiful(?x):-stupendous(?x).
\item beautiful(Mary).
\item @r2 neg stupendous(Mary).
\end{enumerate}
\noindent
The above rules entail that Mary is cute. If later, the speaker says that ``in fact, Mary is also stupendous,'' the strong fact, stupendous(Mary), will be added to the knowledge base and refute rule $r2$.

In addition to the implicature scales related to quantifiers and predicates, Wainer also mentions the implicature scale, $<$or,\ all$>$. For example, the sentence, ``Mary or Tom passed the exam,'' indicates that one of Mary or Tom passes the exam, but the speaker does not know exactly whom. The sentence also implies that ``Mary and Tom pass the exam'' is false by default. For our current approach, we do not support this kind of implicature scale. Instead, we propose to simulate the logical operations \textit{inclusive\ or} and \textit{exclusive\ or} in CNL. We use the connective \textit{or},  to denote the \textit{inclusive\ or}. For example, that sentence, ``John votes for Obama or Romney.'' indicates that John votes at least for one of them. The sentence is formalized in LPDA as shown below:

\begin{enumerate}[label={},nolistsep]
\item vote(John,Romney):-neg vote(John, Obama).
\item vote(John,Obama):-neg vote(John, Romney).
\end{enumerate}
\noindent
The above rules guarantee that if there is a fact that John does not vote Obama (or Romney), then the conclusion that John votes Romney (or Obama) can be derived. Besides, these rules are consistent also when there are the facts that Tom votes for both Obama and Romney. However, if Tom votes neither for Obama nor Romney, it will cause an inconsistency because it violates the constraint that John votes for at least of the candidates. We can use the connective \textit{either \ldots or} to denote the \textit{exclusive\ or}. For example, the sentence, ``John votes either for Obama or for Romney,'' is formalized as 
\begin{enumerate}[label={},nolistsep]
\item neg vote(John,Romney):-vote(John, Obama).
\item neg vote(John,Obama):-vote(John, Romney).
\end{enumerate}
\noindent
If there is a fact that John votes Obama (or Romney), the above rules ensure that the conclusion that John does not vote for Romney (or Obama) can be derived. If John votes for both Obama and Romney, it will cause an inconsistency because it violates the constraint that John can vote for only one of the candidates. 

\section{Conclusion}
In this report, we study controlled natural languages and propose their extensions to support nonmonotonic reasoning. First, we give an overview of CNLs in terms of language design, classifications and their development. Then, we introduce three CNL systems:  Attempto Controlled English (ACE), Processable English (PENG), and Computer-processable English (CPL). We identify their shared traits and distinguishing characteristics in language design, semantic interpretations, and reasoning services. Finally, we propose our extensions of CNL to support defeasible reasoning. We present the representation of \textit{defaults}, \textit{exceptions} and \textit{conversational\ implicatures} in CNL and their formalizations in \textit{Logic\ Programming\ with\ Defaults\ and\ Argumentation\ Theory}.

\newpage
\addcontentsline{toc}{section}{References}
\bibliographystyle{plain}
\bibliography{main}
\appendix
\section{Appendix}
In this section, we introduce three nonmonotonic reasoning frameworks: circumscription, answer set programming and logic programming with defaults and argumentation theories.
\subsection{Circumscription}
Circumscription was first proposed by McCarthy \cite{mccarthy1980circumscription} and later enhanced by Lifschitz \cite{lifschitz1996circumscription} with some extensions to the original approach. Circumscription has  the same syntax as classical logic but with different semantics. Unlike classical logic, circumscription only considers the \textit{preferred} models that have the minimal extensions for certain predicates. For instance, the sentence, ``Eagles are birds. Normally, birds fly,'' is represented as
\begin{equation*}
bird(eagle) \wedge (bird(X) \wedge \neg ab(X) \rightarrow fly(X))
\end{equation*}
\noindent
where $ab$ denotes abnormality. Intuitively, circumscription attempts to minimize the set of objects that are considered to be abnormal and therefore deems an object to be abnormal only if is necessary. For the above formula, since there is no information on the abnormality of eagles, we can assume that eagles are not abnormal and therefore get a conclusion that eagles fly. Formally, let $A(P)$ be a first-order logic sentence containing predicate $P$, the circumscription of $P$ in $A(P)$, denoted as $CIRC[A(P);P]$, is represented as the second-order formula $A^{*}(P)$:
\begin{equation*}
A(P)\wedge\neg\exists p[A(p)\wedge p < P]
\end{equation*}
\noindent
where $p$ is a predicate variable that has the same arity as $P$ and $p<P$ indicates that the extension of $p$ is a strict subset of $P$. The formula says that there does not exist any predicate $p$ such that \romannumeral1) $p$ satisfies all of the conditions of $P$ in $A(P)$ and \romannumeral2) $p$ has a smaller extension than $P$. For the semantics of circumscription, model $M_{1}$ is a submodel of $M_{2}$ in $P$, denoted as $M_{1}\le_{P}M_{2}$, if  \romannumeral1) $M_{1}$ and $M_{2}$ have the same domain, \romannumeral2) the extension of $P$ in $M_{1}$ is a subset of that in $M_{2}$, \romannumeral3) all other predicate and function symbols that occur in A have the same extensions in $M_{1}$ and $M_{2}$. Thus, model $M$ is the minimal model of $A(P)$ with respect to  $P$ if and only if for any model $M'$ of $A(P)$ such that $M'\le_{P}M$,  it implies that $M'=M$. Circumscription leads to a nonmonotonic inference relation: $\nc{A(P)}{\varphi}\ $ iff $A^{*}(P)\models\varphi$. That is, formula $\varphi$ can be inferred from $A(P)$ if and only if $\varphi$ is entailed by $A^{*}(P)$. Extensions of circumscription include parallel circumscription, prioritized circumscription, etc. Details can be found in \cite{lifschitz1996circumscription}.
\subsection{Answer Set Programming}
Answer set programming is based on stable model semantics \cite{gelfond1988stable}. In normal programs, a rule is of the form
\begin{equation*}
c \leftarrow a_{1}, a_{2}, \ldots, a_{m}, not\ b_{1}, not\ b_{2}, \ldots, not\ b_{n}
\end{equation*}
\noindent
where $a_{i}$ ($1\le i \le m$), $b_{j}$ ($1\le j \le n$), and $c$ are atoms. Given an interpretation $M$ of a normal program $\Pi$, the reduct of the program with respect to $M$, denoted as $\Pi^{M}$, is obtained by \romannumeral1) removing the rules that contain negative literals, $not\ l$, where $l\in M$, and \romannumeral2) removing all negative literals that occur in the rest of the rules. The interpretation, $M$, is a stable model of $\Pi$ if $M=LM(\Pi^{M})$ where $LM(\Pi^{M})$ stands for the \textit{least\ model} of $\Pi^{M}$. Stable models are not necessarily unique. There can be zero, one, or multiple stable models for a normal logic program. For example, given the program, $\{q\leftarrow not\ p,\ p\leftarrow not\ q\}$, there are two stable models: $\{p\}$ and $\{q\}$. However, for the program, $\{q\leftarrow not\ p\}$, there is no stable model. 

The formal language used in answer set programming paradigm is called \textit{AnsProlog}, which extends normal programs with \romannumeral1) contraints, \romannumeral2) strong negation, and \romannumeral3) disjunction. An ASP rule is formally represented as
\begin{equation*}
c_{1} \vee c_{2} \vee \cdots \vee c_{k} \leftarrow a_{1}, a_{2}, \ldots, a_{m}, not\ b_{1}, not\ b_{2}, \ldots, not\ b_{n}
\end{equation*}
\noindent
where $a_{i}$ ($1\le i \le m$), $b_{j}$ ($1\le j \le n$), $c_{l}$ ($1\le l \le k$) are literals, $not$ denotes \textit{negation\ as\ failure}, and the rule head is a disjunction of literals. An ASP rule is called a constraint if the rule head is empty. An example is shown below:
\begin{equation*}
\leftarrow l_{1},l_{2},\ldots,l_{n} 
\end{equation*}
\noindent
The constraint forces $l_{i}$ ($1\le i \le n$) not to be included in the models (\textit{answer\ sets}) of the program. 

For the semantics of ASP, let $\Pi_{1}$ be an ASP program that does not contain any negative literal in the body of the rules and $M_{1}$ be a set of ground literals, $M_{1}$ is an \textit{answer\ set} for $\Pi_{1}$ if \romannumeral1) $M_{1}$ satisfies all the rules in $\Pi_{1}$ and \romannumeral2) $M_{1}$ is minimal. Next, let $\Pi_{2}$ be an ASP program that contains negative literals in the body of the rules and $M_{2}$ be a set of ground literals, $M_{2}$ is an \textit{answer\ set} for $\Pi_{2}$ if  $M_{2}$ is the \textit{answer\ set} of $\Pi_{2}^{M2}$, the \textit{reduct} of $\Pi_{2}$ with respect to $M_{2}$. Given the definition of \textit{answer\ sets}, an ASP program $\Pi'$ entails a ground literal $l$ if $l$ is in every \textit{answer\ set} of $\Pi'$. For any query of the form $l_{1} \wedge l_{2} \ldots \wedge l_{n}$ where $l_{i}$ ($1\le i \le n$) is a literal, it will give one of the following answers:
\begin{itemize}
  \item \textit{yes}, if $\Pi \models \{l_{1}, l_{2}, \ldots, l_{n}\}$
  \item \textit{no}, if  there exists $l_{i}$ such that $\Pi \models \neg l_{i}$
  \item \textit{unknown}, otherwise
\end{itemize}

Answer set programming can represent \textit{defaults} and \textit{exceptions}. The general form of a \textit{default} is  
\begin{equation*}
p(X)\leftarrow c(X), not\ ab(d(X)), not\ \neg p(X).
\end{equation*}
\noindent
where $d$ stands for default, $ab$ represents abnormality, $not$ refers to \textit{negation\ as\ failure}, and $\neg$ denotes strong negation. The formula says that ``Normally elements of class $C$ has the property $P$'' or ``$X$ has the property $P$ if 1) $X$ is a class of $C$, 2) $X$ is not abnormal with respect to $d$, and 3) it cannot be shown that $X$ does not have property $P$.'' A \textit{default} can have two types of \textit{exceptions}: \textit{strong\ exception} and \textit{weak\ exception}. A \textit{strong\ exception} refutes the default conclusion and derives the opposite one. A \textit{weak\ exception} stops the application of the \textit{default} without defeating the default conclusion. Let $e(X)$ be a strong \textit{exception} to the \textit{default}. It is represented as
\begin{equation*}
\neg p(X)\leftarrow e(X).
\end{equation*}
\noindent
where $P(X)$ denotes the \textit{default} conclusion. A \textit{weak\ exception} is denoted by a \textit{cancellation\ axiom}. Let $e'(X)$ be a \textit{weak\ exception}, it is represented as 
\begin{equation*}
ab(d(X))\leftarrow not\ \neg e'(X)
\end{equation*}
\noindent
The formula says that $X$ is abnormal with respect to $d$ if there is no evidence that $X$ is not an \textit{exception}. 

\subsection{Logic Programming with Defaults and Argumentation Theories}
Logic programming with defaults and argumentation theories (LPDA) is a unifying defeasible reasoning framework that works on \textit{defaults} and \textit{exceptions} with prioritized rules and argumentation theories. It is based on well-founded semantics \cite{przymusinski1994well}, which are three-valued semantics containing \textit{true}, \textit{false}, and \textit{undefined}. A literal has one of the following forms:
\begin{itemize}
  \item An atomic formula.
  \item neg A, where A is an atomic formula.
  \item not A, where A is an atom.
  \item not neg A, where A is an atom.
  \item not not L and neg neg L, where L is a literal.
\end{itemize}
\noindent
Let $A$ be an atom. A \text{not-free} literal refers to a literal that can be reduced to $A$ or $\neg A$. A \textit{not}-literal refers to a literal that can be reduced to $not\ A$ or $not\ \neg A$. LPDA has two types of rules: \textit{definite} rules and \textit{defeasible} rules. A definite rule is of the form:
\begin{equation*}
L\leftarrow Body
\end{equation*}
\noindent
where $L$ is a \textit{not-free} literal and $Body$ is a conjunction of literals. A defeasible rule is of the form:
\begin{equation*}
@r\ L\leftarrow Body
\end{equation*}
\noindent
where  $r$ is a \textit{term} that denotes the label of the rule. 

Each LPAD program is accompanied with an argumentation theory that decides when a defeasible rule is defeated. An argumentation theory is a set of definite rules with four special predicates: \textbf{defeated}, \textbf{opposes}, \textbf{overrides}, and \textbf{cancel} where \textbf{defeated} denotes the defeatedness of a defeasible rule, \textbf{opposes} indicates the literals that are incompatible with each other,  \textbf{overrides} denotes a binary relation between defeasible rules indicating priority, and \textbf{cancel} cancels a defeasible rule. There are multiple sets of argumentation theories. Users can specify one of them for use or manipulate with the existing ones. A rule is defeated if it is \textit{refuted}, \textit{rebutted}, or \textit{disqualified}. The meaning of \textit{refuted}, \textit{rebutted}, and \textit{disqualified} depends on the argumentation theories. Generally, a rule is \textit{refuted} if there is another rule that draws an incompatible conclusion with a higher priority. A rule is \textit{rebutted} if there is another rule that draws an incompatible conclusion without any priority information. A rule is disqualified if it is \textit{cancelled}, self-defeated, etc.. 

Given an LPDA program $P$ and the argumentation theory $AT$, the well-founded model of $P\cup AT$, denoted as $WFM(P,AT)$, is defined as the limit of the following transfinite induction:
\begin{itemize}
  \item $I_{k} = LPM(\frac{P\cup AT}{I_{k-1}})$, where $k$ is a non-limit ordinal
  \item $I_{k} = \cup_{i\le k}I_{i}$, where $k$ is a limit ordinal
\end{itemize}
\noindent
$P\cup AT$ can be reduced to a normal logic program $P'\cup AT$ such that $P'\cup AT$ has the same well-founded model as $P\cup AT$. The reduction is done by changing every defeasible rule $@r\ L\leftarrow Body$ to a definite rule of the form:
\begin{equation*}
@r\ L \leftarrow Body,\ not\ \$defeated(handle(r,L)) 
\end{equation*}
\noindent
where the term $handle(r,L)$ is called the \textit{handle} of the rule.

\section{Appendix}
In this section, we show the ASP programs for the marathon puzzle and the jobs puzzle.
\subsection{The Marathon Puzzle}
\begin{enumerate}[label={},nolistsep]
\item runner(dominique).
\item runner(ignace).
\item runner(naren).
\item runner(olivier).
\item runner(pascal).
\item runner(philippe).
\item position(1..6).
\item 1 \{ allocated to(A,B) : position(B) \} 1 :- runner(A).
\item :- runner(C), position(D), allocated to(C,D), runner(E), allocated to(E,D), C != E.
\item :- allocated to(olivier,6).
\item before(F,G) :- runner(F), position(H), allocated to(F,H), runner(G), position(I), allocated to(G,I), H < I.
\item :- before(naren,dominique).
\item :- before(naren,pascal).
\item :- before(naren,ignace).
\item :- before(olivier,dominique).
\item :- before(olivier,pascal).
\item :- before(olivier,ignace).
\item :- position(J), J >= 3, allocated to(dominique,J).
\item :- position(K), K > 4, allocated to(philippe,K).
\item :- allocated to(ignace,2).
\item :- allocated to(ignace,3).
\item :- position(L), allocated to(pascal,L), position(M), allocated to(naren,M), L != M - 3.
\item :- allocated to(ignace,4).
\item :- allocated to(dominique,4).
\item answer(N,O) :- runner(N), position(O), allocated to(N,O).
\item \#hide. \#show answer/2.
\end{enumerate}
\subsection{The Jobs Puzzle}
\begin{enumerate}[label={},nolistsep]
\item person(roberta). person(thelma). person(steve). person(pete).
\item female(roberta). female(thelma).
\item male(steve). male(pete).
\item :- person(X), male(X), female(X).
\item 1 {hold(X,Y) : person(X)} 1 :- job(Y).
\item 2 {hold(X,Y) : job(Y)} 2 :- person(X).
\item job(chef). job(guard). job(nurse). job(operator). job(police). job(teacher). job(actor). job(boxer).
\item male(X) :- person(X), job(nurse), hold(X,nurse).
\item male(X) :- person(X), job(actor), hold(X,actor).
\item husband(Y,X) :- person(X), job(chef), hold(X,chef), person(Y), job(operator), hold(Y,operator).
\item male(X) :- person(X), person(Y), husband(X,Y).
\item female(Y) :- person(X), person(Y), husband(X,Y).
\item :- job(boxer), hold(roberta,boxer).
\item :- educated(pete).
\item educated(X) :- person(X), job(nurse), hold(X,nurse).
\item educated(X) :- person(X), job(police), hold(X,police).
\item educated(X) :- person(X), job(teacher), hold(X,teacher).
\item :- job(chef), hold(roberta,chef).
\item :- job(police), hold(roberta,police).
\item :- person(X), job(chef), hold(X,chef), job(police),hold(X,police).
\item answer(hold(X,Y)) :- job(Y), hold(X,Y).
\end{enumerate}

\end{document}